\documentclass[lettersize,journal]{IEEEtran}
\usepackage{amsmath,amsfonts}
\usepackage{array}
\usepackage[ruled,vlined]{algorithm2e}
\usepackage[caption=false,font=normalsize,labelfont=sf,textfont=sf]{subfig}
\usepackage[pagebackref,breaklinks,colorlinks]{hyperref}
\usepackage{textcomp}
\usepackage{stfloats}
\usepackage{url}
\usepackage{bm}
\usepackage{verbatim}
\usepackage{graphicx}
\usepackage{booktabs}   
\usepackage{colortbl}   
\usepackage{xcolor}
\usepackage{mathtools}
\usepackage{multirow}
\usepackage{framed}
\usepackage{cite}
\usepackage{bbding}
\usepackage{makecell}
\usepackage{placeins}
\definecolor{oursbg}{RGB}{226,235,230}
\newcommand{\model}{ESIA}
\usepackage{fancyhdr}
\fancypagestyle{firstpage}{
  \fancyhead{} 
  \fancyhead[C]{ \textit{Accepted at IEEE Transactions on Multimedia.}}
  }

\begin{document}

\title{ESIA: An Energy-Based Spatiotemporal Interaction-Aware Framework for Pedestrian Intention Prediction}
\author{Yanping Wu,
        Meiting Dang,
        Lin Wu,
        Edmond S. L. Ho,
        Zhenghua Chen,
        Chongfeng Wei\IEEEauthorrefmark{1}%
\thanks{\IEEEauthorrefmark{1}Chongfeng Wei is the corresponding author.}
\thanks{Yanping Wu, Meiting Dang, Lin Wu, Zhenghua Chen and Chongfeng Wei are with the James Watt School of Engineering, University of Glasgow, Glasgow, G12 8QQ, United Kingdom (e-mail: 3066431W@student.gla.ac.uk; \{m.dang.1, l.wu.1\}@research.gla.ac.uk; \{Zhenghua.Chen, Chongfeng.Wei\}@glasgow.ac.uk).}
\thanks{Edmond S. L. Ho is with the School of Computing Science, University of Glasgow, Glasgow, G12 8QQ, United Kingdom (e-mail: Shu-Lim.Ho@glasgow.ac.uk).}}




\maketitle
\thispagestyle{firstpage}

\begin{abstract}
Recent advances in autonomous driving have motivated research on pedestrian intention prediction, which aims to infer future crossing decisions and actions by modeling temporal dynamics, social interactions, and environmental context.
However, existing studies remain constrained by oversimplified multi-agent interaction patterns, opaque reasoning logic, and a lack of global consistency in behavioral predictions, which compromise both robustness and interpretability.
In this work, we propose \model\ (\underline{E}nergy-based \underline{S}patiotemporal \underline{I}nteraction-\underline{A}ware framework), a novel Conditional Random Field (CRF)-based paradigm.
We cast the intention prediction task as a structured prediction problem over a unified graph-based representation, treating pedestrians and the environment as spatiotemporal nodes.
To characterize their distinct roles, we assign unary potentials to nodes to capture individual intentions, and pairwise potentials to edges to encode social and environmental interactions.
These potentials are integrated into a unified global energy function to ensure scene-level consistency across behavioral predictions.
To further constrain inference without ground-truth supervision, we introduce structural consistency terms to penalize logical contradictions.
This optimization is efficiently solved via a novel Unary-Seeded Simulated Annealing (U-SSA) algorithm, which leverages high-confidence unary priors to rapidly converge to a high-quality solution.
Extensive experiments on standard benchmarks demonstrate that \model\ achieves state-of-the-art performance with improved interpretability over existing methods.
\end{abstract}

\section{Introduction}
\IEEEPARstart{P}{edestrian} intention prediction has witnessed remarkable progress in autonomous driving, showcasing its potential for proactive risk mitigation~\cite{9743954,chen2021visual,perdana2021early}.
However, real-world traffic scenarios involve diverse categories of agents---pedestrians, cyclists, vehicles, and road infrastructure~\cite{caesar2020nuscenes,mei2022waymo}---whose behaviors are governed by heterogeneous dynamics~\cite{ma2019trafficpredict} and social conventions~\cite{helbing1995social,alahi2016social}. As a result, multi-agent interactions are inherently complex, highly context-dependent, and difficult to predict, particularly when agents exhibit ambiguous or abrupt behavioral transitions under dense traffic conditions~\cite{zhao2024autonomous,hu2023planning}.

To address these challenges, existing studies have progressively integrated pedestrian intention prediction into the perception, decision-making, and planning pipelines of autonomous driving systems~\cite{dang2024coupling,zhou2023pit,kotseruba2021benchmark,gesnouin2021trouspi,chen2024pedestrian}.
Early works relied on handcrafted features and rule-based models~\cite{kooij2014context,schneider2013pedestrian}, while recent deep learning approaches leverage recurrent neural networks~\cite{alahi2016social}, graph-based representations~\cite{mohamed2020social,cadena2022pedestrian}, and attention mechanisms~\cite{yuan2021agentformer,zhou2023pit} to capture spatio-temporal dependencies. These methods have demonstrated promising results in benchmarks~\cite{rasouli2017ICCVW,rasouli2019pie}, validating their effectiveness in system safety and reliability.

Despite their significant success, current approaches for pedestrian intention prediction remain bottlenecked by opaque and rigid interaction representations, suffering from the following three key limitations:
\textbf{1) Inadequate Interaction Modeling:} 
Real-world decision-making is driven by the highly dynamic interplay between intrinsic intentions and extrinsic environmental constraints. Yet, existing paradigms often oversimplify these relationships by treating agents in isolation or ignoring cross-modal dependencies, failing to capture the diversity and complexity of real-world interaction patterns.
\textbf{2) Limited Interpretability:} The failure to explicitly model individual intentions and multi-dimensional interactions renders the decision rationale opaque, which significantly undermines the trustworthiness of prediction results for downstream tasks. In other words, the learned models cannot provide explicit reasoning such as \textit{``The pedestrian intends to cross due to the approaching vehicle's deceleration''}.
\textbf{3) Global Inconsistency:} Existing methods often obscure the semantic coupling between heterogeneous features and lack a unified optimization objective. Consequently, they may fail to guarantee global consistency, resulting in predictions that appear locally reasonable but are globally physically inconsistent.

\begin{figure}[t]
    \centering
    \subfloat[Traditional Methods.]{%
        \label{fig:introd_a}%
        \includegraphics[height=4.5cm]{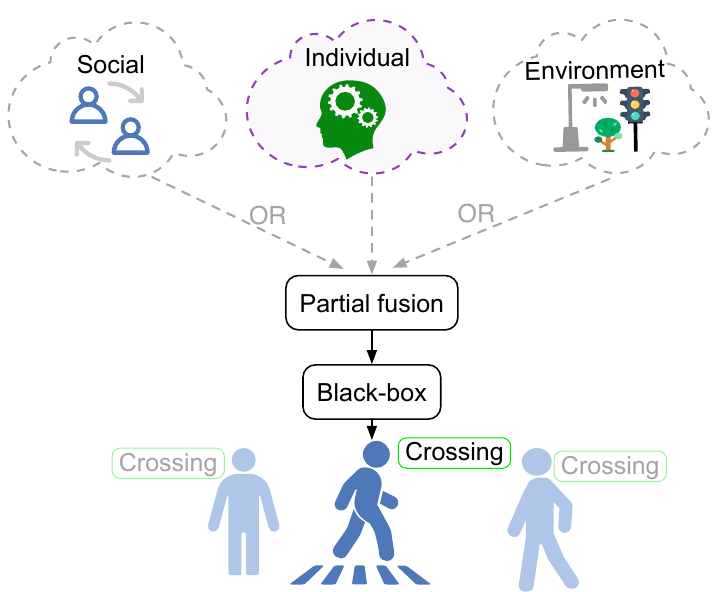}%
    }
    \subfloat[Our \model.]{%
        \label{fig:introd_b}%
        \includegraphics[height=4.5cm]{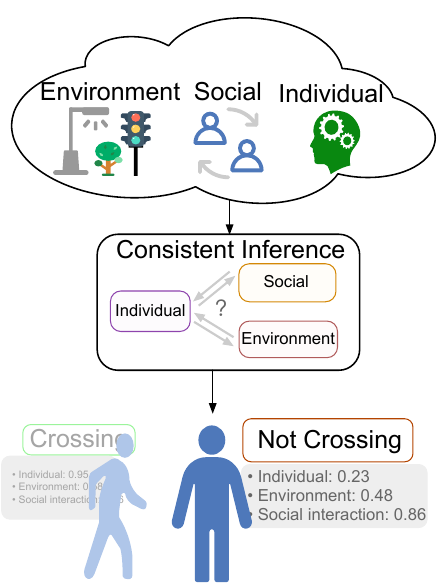}%
    }
    \caption{Illustration of paradigms for pedestrian intention prediction. (a) Traditional methods often employ naive fusion within black-box architectures, leading to opaque reasoning and limited reliability. (b) The \model\ explicitly captures pedestrian--pedestrian interactions, environmental constraints, and individual intentions. By enforcing structural consistency during inference, \model\ achieves both robust performance and superior interpretability.}
    \label{fig:introd}
\end{figure}

These limitations pose significant challenges to achieving interpretable predictions in dynamic traffic scenarios. Specifically, an ideal system requires multi-dimensional interaction modeling, transparent reasoning, and physically plausible global consistency.
However, existing methods tend to tackle these challenges in isolation, as illustrated in Fig.~\ref{fig:introd_a}.
For instance, some studies focus exclusively on pedestrian-centric features or social interactions~\cite{xie2024pedestrian,sofianos2021space,cadena2022pedestrian}, without accounting for the fact that pedestrian behaviors are shaped by the complex interplay of intrinsic states, social dynamics, and environmental constraints.
Other methods operate as opaque \textit{black-box} models with limited interpretability~\cite{xie2025gtranspdm,zhou2023pit,chen2024pedestrian},
thus obscuring the underlying decision rationale and making predictions difficult to trust.
More importantly, many approaches fail to guarantee global consistency with respect to physical and social constraints~\cite{zhang2023trep}, resulting in a fragmented treatment of different factors that fails to preserve the inherent coupling. For example, a pedestrian's decision reflects a trade-off between ``\textit{yielding to vehicles}'' and ``\textit{following the crowd}''. 
This raises a fundamental yet underexplored question: 
\begin{framed}
\textit{Can we achieve multi-agent interaction modeling, interpretable reasoning, and globally rational pedestrian intention prediction within a unified framework?}
\end{framed}

To answer the above question, we propose an \textbf{E}nergy-based \textbf{S}patiotemporal \textbf{I}nteraction-\textbf{A}ware (\textit{\textbf{\model}}) framework (see Fig.~\ref{fig:introd_b}), which is designed to jointly achieve complex interaction modeling and interpretable, socially consistent prediction.
\textbf{First,} we revisit the pedestrian intention prediction problem from a multi-level interaction perspective. 
Intuitively, crossing behavior fundamentally arises from the coupling of three factors: individual states (e.g., \textit{head orientation and body pose}), social interactions (e.g., \textit{group motion and avoidance patterns}), and environmental constraints (e.g., \textit{traffic signals and vehicle dynamics}). 
Achieving consistent decisions across these three dimensions is crucial for accurate and interpretable predictions.
\textbf{Second,} to explicitly enhance interpretability in pedestrian intention prediction, we reformulate the interaction paradigm under a unified framework. Inspired by the Conditional Random Field (CRF) paradigm, we cast individual intentions as unary potentials while formulating pedestrian-pedestrian (P--P) and pedestrian-environment (P--E) interactions as 
pairwise potentials. This establishes a paradigm shift from implicit black-box representations to an explicit interaction-aware formulation.
\textbf{Third,} we integrate unary and pairwise potentials into a unified global energy framework to capture physical and social constraints. Notably, to resolve logical contradictions in the absence of explicit supervision, we introduce structural consistency energy that penalizes discrepancies between predictions. Furthermore, \model\ decouples feature extraction from interaction modeling, enabling plug-and-play extensibility.

Our main contributions are as follows:
\begin{itemize}
    \item  
    We propose \model, which reformulates pedestrian intention prediction into a structured energy minimization problem.
    By explicitly decoupling individual goals, social interactions, and environmental constraints into unary and pairwise potentials, \model\ provides interpretable reasoning and a flexible plug-and-play architecture.
    \item 
    We introduce structural consistency energy terms to penalize logical contradictions during inference without ground-truth supervision. To achieve global consistency, we propose a Unary-Seeded Simulated Annealing (U-SSA) strategy that leverages high-confidence unary priors as search seeds to efficiently navigate the energy landscape and find a high-quality configuration.
    \item 
  We conduct extensive experiments on the JAAD and PIE datasets, demonstrating that \model\ achieves state-of-the-art performance with enhanced interpretability over existing methods.
\end{itemize}

\section{Related Work}
\label{sec:rw}
In this section, we review prior work related to our study, including pedestrian intention prediction, interaction-aware pedestrian intention prediction, and CRF-based methods.

\textbf{Pedestrian Intention Prediction.}
Pedestrians' intentions can be accurately inferred by jointly exploiting rich cues, such as individual pedestrian states, social interactions, and environmental context~\cite{azarmi2025pedestrian,7759351,7313340}.
Anticipating such intentions is pivotal for enabling safer and more robust decision-making in traffic. 
Existing methods can be categorized into three paradigms according to their modeling strategies.
First, physics- and kinematics-based methods are computationally efficient but struggle to handle complex scenarios~\cite{scholler2020constant,helbing1995social}. Second, traditional machine learning-based approaches significantly enhance flexibility in learning nonlinear motion~\cite{kooij2014context,yau2021graph}.
Third, the dominant deep learning-based methods utilize powerful representation capabilities to achieve precise and robust predictions~\cite{zhangji2025reliable,dong2024sparse,yang2020social,zhangji2023channel}.

\textbf{Interaction-Aware Pedestrian Intention Prediction.} Interaction modeling aims to quantify the complex influences in traffic scenes, including P--P social interactions (\textit{e.g., collision avoidance and group walking}) and P--E interactions (\textit{e.g., traffic signals})~\cite{10899874}.
Existing methods employ GCNs to capture spatial correlations~\cite{mohamed2020social,shi2021sgcn} or attention mechanisms to model temporal dependencies~\cite{yu2020spatio,yuan2021agentformer}, with some integrating semantic map cues~\cite{zhou2023query}. Despite these advances, a fundamental limitation persists: these methods largely operate as black-box models. Specifically, GCNs typically treat edges as simple aggregation pathways rather than explicit physical constraints, limiting interpretability. In contrast, we reformulate interactions as explicit energy potentials within the CRF paradigm, enabling interpretable reasoning and ensuring global consistency via energy minimization.

\textbf{CRF-based Methods.} Conditional Random Fields (CRFs) model the conditional probability distribution over output variables given observations~\cite{zheng2015conditional,hou2022detecting,zhang2024msd,mahendran2024novel}.
They have been extensively applied in pixel-level tasks like semantic segmentation and depth estimation~\cite{zhang2024msd,wang2024multimodal,rakic2024tyche,naseer2024efficient}, as well as for modeling sequential constraints in natural language processing~\cite{wu2024adaptive,Arslan2024application} and dynamic systems in vehicular networks~\cite{mahendran2024novel}. However, their potential for pedestrian intention prediction remains largely unexplored. To the best of our knowledge, this is the first work to introduce CRFs to model pedestrian intention consistency in complex traffic scenarios.

\section{Preliminaries}
\label{sec:pre}
\subsection{Conditional Random Field (CRF)}
\label{sec:crf_definition}
We model pedestrian intention prediction as a labeling problem on an undirected graph $\mathcal{G} = (\mathcal{V}, \mathcal{E})$, where the node set $\mathcal{V}$ represents pedestrians and the edge set $\mathcal{E}$ denotes interactions. 
Let $\boldsymbol{o}$ be the observations and $\boldsymbol{y}$ be the label configuration. The conditional probability $P(\boldsymbol{y} \mid \boldsymbol{o})$ follows the Gibbs distribution~\cite{li2022comprehensive,vanmarcke2010random}:
\begin{equation}
P(\boldsymbol{y} \mid \boldsymbol{o};\boldsymbol{\theta}) = \frac{1}{Z(\boldsymbol{o};\boldsymbol{\theta})} \exp\big(-E(\boldsymbol{y},\boldsymbol{o};\boldsymbol{\theta})\big),
\label{eq:crf_gibbs}
\end{equation}
where $Z(\cdot)$ is the partition function and $E(\cdot)$ is the global energy.
To balance interpretability and computational efficiency, particularly in sparse real-world scenes, we restrict the energy function to unary (node) and pairwise (edge) terms, omitting higher-order cliques, as they introduce exponential computational overhead without substantial gains in our setting.
Thus, the total energy decomposes as:
\begin{equation}
E(\boldsymbol{y},\boldsymbol{o};\boldsymbol{\theta}) = \sum_{i \in \mathcal{V}} \Phi(y_i, \boldsymbol{o}_i;\boldsymbol{\theta}) + \sum_{(i,j) \in \mathcal{E}} \Psi(y_i, y_j, \boldsymbol{o}_{ij};\boldsymbol{\theta}).
\end{equation}
This decomposition captures the intuition that a pedestrian's intention is driven by individual cues and pairwise interactions within a local neighborhood.

\subsection{Problem Formulation}
Given a sequence of consecutive frames $\boldsymbol{f}=[\boldsymbol{f}_0,\boldsymbol{f}_1,\dots,\boldsymbol{f}_{T-1}]$, each frame 
$\boldsymbol{f}_t \in \mathbb{R}^{H\times W \times C}$ represents a front-view image
captured by the ego vehicle, where $H$ and $W$ denote the image height and width, and $C$ is the number of channels. 
At time step $t$, we extract $l$ consecutive observation frames, denoted as $\boldsymbol{\mathcal{F}}^{(t)} = [\boldsymbol{f}_{t-l+1}, \dots, \boldsymbol{f}_t]$.
Specifically, $\boldsymbol{\mathcal{F}}^{(t)}$ 
contains observations of $n$ pedestrians tracked consistently across these $l$ frames. The input representations are formulated as follows.

\textbf{Pedestrians.} Assume that $\boldsymbol{f}_t$ contains $n$ pedestrians. For the $i$-th pedestrian $x_{i}$ in $\boldsymbol{f}_t$, the bounding box coordinates are denoted as $\boldsymbol{C}_i^t=(x_\mathtt{min}^{t,i},y_\mathtt{min}^{t,i},x_\mathtt{max}^{t,i},y_\mathtt{max}^{t,i})$, and the corresponding cropped image is represented as $\boldsymbol{I}_i^t$.
For each pedestrian $x_i$ (where $i \in \{1,\dots, n\}$) extracted from $\boldsymbol{\mathcal{F}}^{(t)}$, we adopt the following notations:
\begin{equation}
\begin{aligned}
\boldsymbol{I}_i &= [\boldsymbol{I}_i^{t-l+1},\dots,\boldsymbol{I}_i^{t}], \\
\boldsymbol{C}_i &= [\boldsymbol{C}_i^{t-l+1},\dots,\boldsymbol{C}_i^{t}].
\end{aligned}
\end{equation}
Consequently, the cropped images and bounding boxes for all $n$ pedestrians 
across frames $t-l+1$ to $t$ are denoted as $\boldsymbol{I} = [\boldsymbol{I}_1, \dots, \boldsymbol{I}_{n}]$ and 
$\boldsymbol{C} = [\boldsymbol{C}_1, \dots, \boldsymbol{C}_{n}]$, respectively.

\textbf{Environment.}
We utilize the full frame sequence $\boldsymbol{\mathcal{F}}^{(t)}$ to capture the global environmental context.
Additionally, the ego vehicle speed sequence is defined as $\boldsymbol{V}=[v_{t - l+1}, \dots, v_t]$.
Let $y_i \in \{0, 1\}$ denote the ground-truth intention label for the $i$-th pedestrian, where $y_i=1$ indicates crossing and $y_i=0$ indicates not crossing.
Our goal is to estimate the crossing intention of all $n$ pedestrians over a future time horizon of $1\text{--}2$~s, given the observation window of $l$ frames ($\approx 0.5$~s). Formally, the intention prediction is formulated as:
\begin{equation}
    \hat{p}_i = P(y_i=1 \mid \boldsymbol{I}, \boldsymbol{C}, \boldsymbol{\mathcal{F}}^{(t)}, \boldsymbol{V}),
\end{equation}
where $\hat{p}_i$ denotes the crossing probability of pedestrian $x_i$. 
Here, $\boldsymbol{o} = \{\boldsymbol{I}, \boldsymbol{C}, \boldsymbol{\mathcal{F}}^{(t)}, \boldsymbol{V}\}$ corresponds to the observation defined in Sec.~\ref{sec:crf_definition}, where $\boldsymbol{\mathcal{F}}^{(t)}$ additionally captures global environmental context.

\section{Methodology}
\label{sec:method}
In this section, we first outline the overall framework of \model. We then elaborate on its key technical components, followed by the optimization and inference procedures.

\subsection{Framework Overview}
\begin{figure*}[t] 
    \centering 
    \includegraphics[height=6.1cm,width=18cm]{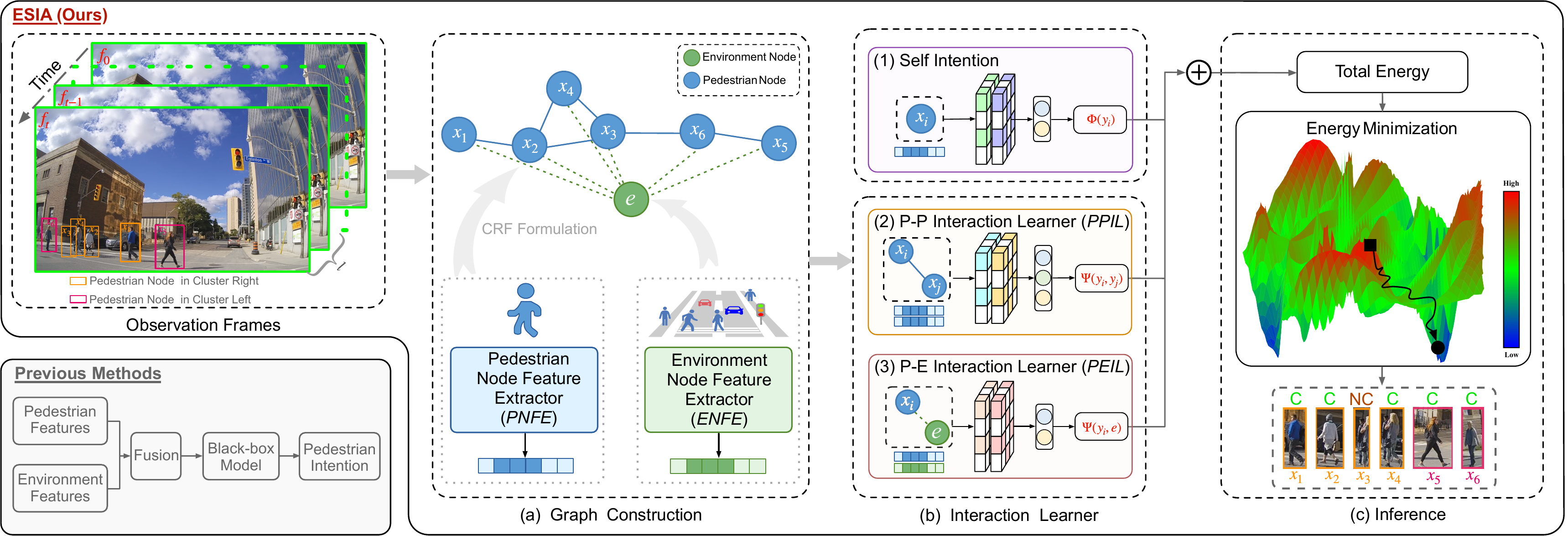} 
    \caption{Overview of our \model. Unlike previous methods (\textit{bottom left}) that treat interaction modeling as a black-box fusion process, our \model\ (\textit{top}) formulates intention prediction within a CRF framework. 
    The \model\ framework consists of three main components: a) Graph Construction, where the scene is modeled as a graph containing pedestrian nodes ($x_i$) and a global environment node ($e$); b) Interaction Learner, which explicitly models three types of potentials: self-intention, P--P interaction, and P--E interaction; and c) Inference, where the intention configuration is derived by minimizing the total energy of the system. Here, \textit{NC} denotes Not Crossing, and \textit{C} denotes Crossing. }
    \label{fig:framework} 
\end{figure*}

Fig.~\ref{fig:framework} illustrates the architecture of \model. Unlike previous black-box approaches, \model\ formulates intention prediction as a structured CRF problem, operating through three sequential stages:
\textbf{a) Construct the scene graph.} The framework first transforms raw observation frames into a structured representation. By extracting pedestrian-level and environment-level features, we establish an undirected graph where pedestrians ($x_i$) and the global environment ($e$) are treated as interacting nodes, which provides the topological foundation for reasoning. \textbf{b) Model interactions via energy potentials.} On the constructed graph, we design an energy function to quantify behavioral compatibility. This stage learns unary potentials to capture individual motion trends, as well as distinct pairwise potentials to encode P--P and P--E interactions. This mechanism encodes abstract spatiotemporal dependencies into measurable energy values. \textbf{c) Infer intentions via global energy minimization.} The framework seeks the optimal intention configuration for the entire scene. Instead of predicting individuals in isolation, we minimize the total energy to jointly estimate the most consistent intentions (\textit{Crossing vs. Not Crossing}) for all pedestrians simultaneously.


\subsection{Graph Construction}
\label{sec:graph}
We formulate the scene as an undirected graph $\mathcal G=(\mathcal{V},\mathcal{E})$, where the node set $\mathcal{V}$ represents scene agents and the edge set $\mathcal{E}$ denotes interactions. Specifically, $\mathcal{V}=\mathcal{V}_\mathtt{p} \cup \{e\}$
comprises pedestrian nodes $\mathcal{V}_\mathtt{p}=\{x_1,\ldots,x_{n}\}$ and a shared environment node $e$. 
The edge set $\mathcal E$ consists of P--P edges $\mathcal E_\mathtt{pp}$ and P--E edges $\mathcal E_\mathtt{pe}$. 
To explicitly model interactions, we cluster pedestrians into left-facing and right-facing groups based on head orientation. 
This design is motivated by the observation that head orientation 
serves as a strong proxy for pedestrian intention~\cite{kooij2014context, schulz2015pedestrian}, and 
pedestrians facing the same direction are more likely to exhibit 
collective crossing behavior, whereas those from opposite sides 
tend to interact as conflicting agents.
The P--P edges $\mathcal{E}_\mathtt{pp}$ are constructed by: 
(1) connecting nodes within the same cluster if their spatial distance $d_{ij} < \delta_d$. The $\delta_d$ is introduced to filter out 
spatially distant pedestrians whose interactions are negligible, 
thereby reducing computational complexity and suppressing noise 
from irrelevant pairings; and (2) linking the closest pair of nodes between distinct clusters. 
As a fallback, if head orientation labels are unavailable, P--P edges are constructed solely based on the threshold $\delta_d$.
Finally, the environment node $e$ connects to all pedestrians ($\mathcal{E}_\mathtt{pe}$). 
Unlike pedestrian nodes, the environment node $e$ 
is not associated with any intention variable; instead, it serves as a global context provider for all pedestrians.

\subsection{Node Feature Encoding and Potentials}
\label{sec:node}
For brevity, we write $\Phi(y_i)$ for the unary potential $\Phi(y_i, \boldsymbol{o}_i;\boldsymbol{\theta})$ and $\Psi(\cdot,\cdot)$ for the pairwise potentials $\Psi(\cdot,\cdot,\boldsymbol{o};\boldsymbol{\theta})$, omitting the explicit dependence on observed features and model parameters.

\begin{figure}[t] 
    \centering 
    \includegraphics[height=5.2cm,width=8cm]{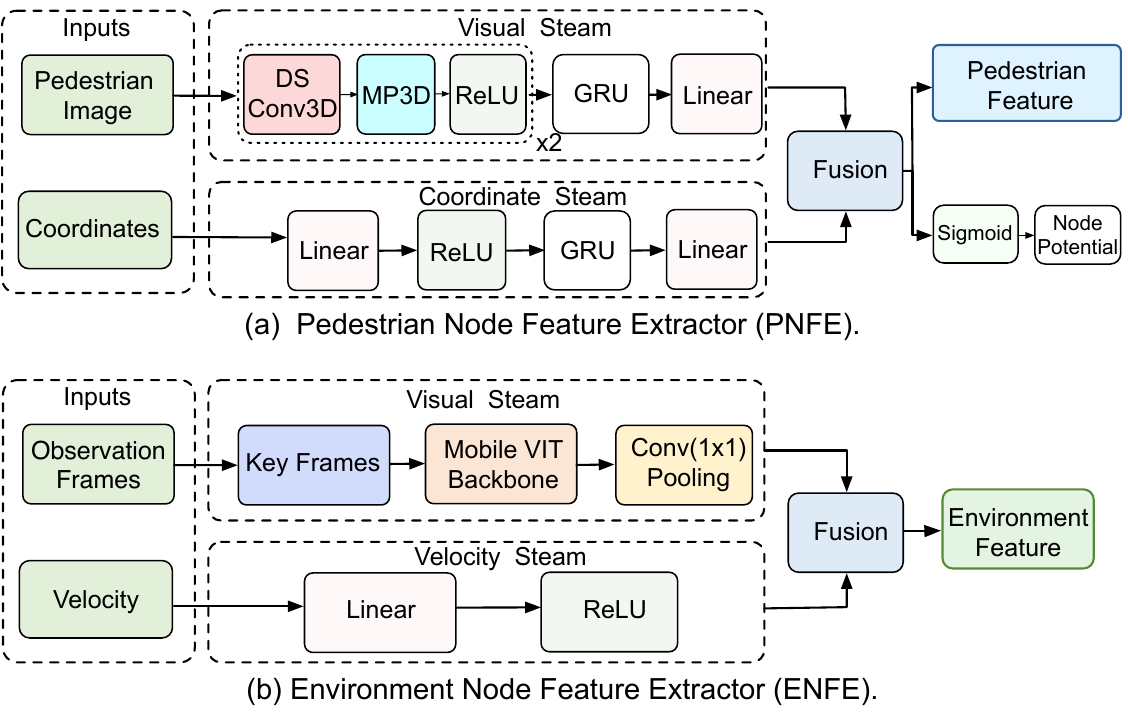} 
    \caption{Architecture of the Feature Extraction Modules. 
    (a) The \textit{PNFE} fuses visual appearance and coordinate cues to generate pedestrian representations and unary potentials. 
    (b) The \textit{ENFE} extracts global context via MobileViT and incorporates velocity to form the environment feature.}
    \label{fig:PNFE} 
\end{figure}

\textbf{Pedestrian Node Feature Extractor (PNFE).}
The \textit{PNFE} module extracts pedestrian-specific features to compute unary potentials. As illustrated in Fig.~\ref{fig:PNFE} (a), we employ a two-stream architecture. The \textit{visual stream} captures spatiotemporal dynamics using a lightweight Depthwise Separable 3D Convolutional Network (DS-Conv3D) followed by 3D max pooling. Specifically, for pedestrian $x_i$, the cropped image sequence $\bm{I}_i$ is processed through stacked blocks of DS-Conv3D, MaxPool3D, and ReLU activation.
The resulting hierarchical features are flattened and then fed into a Gated Recurrent Unit (GRU) followed by a linear layer to capture long-term temporal dependencies, yielding the visual representation $\boldsymbol{h}_x^i$. 
In parallel, the \textit{coordinate stream} encodes spatial trajectory dynamics. It processes the scaled bounding box coordinates $\boldsymbol{C}_i$ through a Linear-ReLU block (a linear layer followed by ReLU activation) and a GRU followed by another linear layer, producing the geometric embedding $\boldsymbol{h}_c^i$.
These two streams are concatenated and passed through a fully connected (FC) layer to predict the crossing probability $\hat{p}_i$:
\begin{equation}
\label{eq:feat_fusion}
\hat{p}_i = \sigma(\text{FC}(\boldsymbol{F}_{x}^{i})), \quad \text{where} \quad \boldsymbol{F}_{x}^{i} = [\boldsymbol{h}_x^i; \boldsymbol{h}_c^i ].
\end{equation}
Here, $\sigma(\cdot)$ denotes the sigmoid activation and $[\cdot;\cdot]$ represents concatenation. 
The unary potential $\Phi(y_i)$ is then defined as the negative log-likelihood of the label assignment $y_i$:
\begin{equation}
\label{eq:node}
\Phi(y_i) = - \left[ y_i \log(\hat{p}_i) + (1 - y_i) \log(1 - \hat{p}_i) \right].
\end{equation}
This term quantifies the intrinsic cost of assigning intention $y_i$ to pedestrian $x_i$, considering only individual-level features.

\textbf{Environment Node Feature Extractor (ENFE).} 
To capture the global scene context and assess crossing feasibility, we introduce a shared environment node $e$. 
As illustrated in Fig.~\ref{fig:PNFE} (b), the \textit{ENFE} module employs a two-stream architecture to encode both static scene context and ego-vehicle velocity. 
The \textit{visual stream} extracts spatial features from the driving environment. 
To reduce computational redundancy, we first select $l_{k}$ key frames based on the Structural Similarity Index Measure (SSIM)~\cite{Mudeng2022prospects}. 
SSIM is adopted because it effectively identifies frames with significant scene changes, thereby retaining the most informative environmental cues while avoiding redundant 
computation from visually similar consecutive frames.
These key frames are processed by a pre-trained MobileViT backbone \cite{mehta2021mobilevit} to capture multi-scale spatial dependencies, with the resulting features pooled into a static environment representation $\boldsymbol{h}_e$. 
Simultaneously, the \textit{velocity stream} encodes the ego-motion via a Linear-ReLU block to obtain the motion embedding $\boldsymbol{h}_v$, capturing dynamic cues that significantly influence pedestrian decision-making.
The environment feature is then formulated by concatenating these embeddings:
\begin{equation}
\label{eq:feat_env}
\boldsymbol{F}_e = [\boldsymbol{h}_e; \boldsymbol{h}_v].
\end{equation}
This feature serves as a unified shared context across all pedestrian nodes,
providing environmental cues for P--E interaction.

\subsection{Edge Potentials}
\label{sec:edge}
\textbf{Pedestrian-Pedestrian Interaction Learner (\textit{PPIL}).}
To capture social norms such as group coherence and conflict avoidance, the \textit{PPIL} models pairwise interactions explicitly.
Instead of simple concatenation, we employ a Multi-Head Attention (MHA) mechanism to enable dynamic information exchange between pedestrian pairs~\cite{vaswani2017attention}.
Given features $\boldsymbol{F}_{x}^{i}$ and $\boldsymbol{F}_{x}^{j}$ from Eq.~\ref{eq:feat_fusion}, we first concatenate them along the sequence dimension to form a pair sequence $\boldsymbol{S}_{ij} \in \mathbb{R}^{2 \times D}$. This sequence is fed into the MHA module to capture mutual dependencies:
\begin{equation}
\label{eq:pp_attn}
\tilde{\boldsymbol{S}}_{ij} = \mathtt{MHA}(\boldsymbol{S}_{ij}),
\end{equation}
where $\mathtt{MHA}(\cdot)$ allows each pedestrian to attend to the other's features.
The refined features $\tilde{\boldsymbol{S}}_{ij}$ are then flattened and passed through a Multilayer Perceptron (MLP) to predict the interaction probability distribution $\hat{\boldsymbol{p}}_{ij} \in \mathbb{R}^3$:
\begin{equation}
\hat{\boldsymbol{p}}_{ij} = \mathtt{Softmax}(\mathtt{MLP}(\mathtt{Flatten}(\tilde{\boldsymbol{S}}_{ij}))).
\end{equation}
Here, the output space corresponds to three interaction states: inconsistent intentions ($k=0$), consistent not-crossing ($k=1$), and consistent crossing ($k=2$).
Accordingly, the pairwise potential $\Psi(y_i, y_j)$ is defined as the negative log-likelihood of the corresponding interaction state:
\begin{equation}
\label{eq:edge_pp}
\Psi(y_i, y_j) = - \log \left( \hat{p}_{ij}^{(k)} \right), \quad \text{where } k = \mathcal{T}(y_i, y_j).
\end{equation}
The mapping $\mathcal{T}$ assigns the label state based on the pair configuration: $\mathcal{T}(0,0)=1$, $\mathcal{T}(1,1)=2$, and $\mathcal{T}(0,1)=\mathcal{T}(1,0)=0$.
This term serves as a soft constraint in the CRF, penalizing intentions that contradict the predicted group dynamics.

\textbf{Pedestrian-Environment Interaction Learner (\textit{PEIL}).}
This module assesses the environmental crossing feasibility to determine whether current conditions (e.g., road geometry, traffic flow) support crossing behavior. 
Given the pedestrian feature $\boldsymbol{F}_{x}^{i}$ (Eq.~\ref{eq:feat_fusion}) and the global environment feature $\boldsymbol{F}_e$ (Eq.~\ref{eq:feat_env}), we concatenate them along the sequence dimension to form a sequence $\boldsymbol{S}_{ie} \in \mathbb{R}^{2 \times D}$. Similar to \textit{PPIL}, we employ an MHA mechanism to model the dependencies between the pedestrian and the scene context:
\begin{equation}
\label{eq:pe_attn}
\hat{p}_{ie} = \sigma(\mathtt{MLP}(\mathtt{Flatten}(\mathtt{MHA}(\boldsymbol{S}_{ie})))),
\end{equation}
where $\hat{p}_{ie}$ represents the predicted probability that the environment facilitates crossing.
The P--E pairwise potential $\Psi(y_i, e)$ is defined as the cost of assigning intention $y_i$ under the observed environmental context:
\begin{equation}
\label{eq:edge_pe}
\Psi(y_i, e) = - \left[ y_i \log(\hat{p}_{ie}) + (1 - y_i) \log(1 - \hat{p}_{ie}) \right].
\end{equation}
This potential ensures that the predicted intentions $y_i$ are consistent with the environmental context. For instance, if the environment is non-supportive ($\hat{p}_{ie} \approx 0$), assigning a crossing intention ($y_i=1$) will incur a high energy penalty.

\subsection{Optimization}
\label{sec:opti}
We formulate the intention prediction task as an energy minimization problem within a CRF framework. The conditional probability of the label configuration $\mathbf{y}$ given the observation $\boldsymbol{o}$ is modeled as:
\begin{equation}
\label{eq:prob-energy}
P(\boldsymbol{y}\mid \boldsymbol{o};\boldsymbol{\theta}) = \frac{1}{Z(\boldsymbol{o};\boldsymbol{\theta})} \exp\left\{-E(\boldsymbol{y},\boldsymbol{o};\boldsymbol{\theta})\right\},
\end{equation}
where $Z$ is the partition function and  $E$ denotes the total energy. The energy decomposes into the weighted sum of unary and pairwise potentials defined in Sec.~\ref{sec:node} and Sec.~\ref{sec:edge}:
\begin{equation}
\begin{aligned}
\label{eq:energy}
E(\boldsymbol{y},\boldsymbol{o};\boldsymbol{\theta}) &= \alpha \sum_{x_i \in \mathcal{V}_\mathtt{p}} \Phi(y_i) + \beta \sum_{(i,j) \in \mathcal{E}_\mathtt{pp}} \Psi(y_i, y_j)\\
&+ \gamma \sum_{(i,e) \in \mathcal{E}_\mathtt{pe}} \Psi(y_i, e).
\end{aligned}
\end{equation}
Here, $\alpha, \beta, \gamma$ are hyperparameters balancing the contributions of individual node features, pedestrian social interactions, and environmental constraints.
To estimate the optimal $\boldsymbol{\theta}$, we minimize the negative log-likelihood of the ground-truth configurations. 
Since each potential is formulated as a cross-entropy loss based on the module's probability output, the training objective for a sample at time $t$ is:
\begin{equation}
\label{equ:total_loss}
\mathcal{L}_t(\boldsymbol{\theta}) = E(\boldsymbol{y}_{\mathtt{gt}},\boldsymbol{o};\boldsymbol{\theta}) + \rho \|\boldsymbol{\theta}\|_2^2,
\end{equation}
where $\boldsymbol{y}_{\mathtt{gt}}$ denotes the ground-truth labels and $\rho$ is the regularization coefficient. 
The model is optimized end-to-end via stochastic gradient descent with backpropagation.
The training pipeline of \model\ is summarized in Algorithm~\ref{algo:train}.

\begin{algorithm}[htbp]
\caption{Training pipeline of \model}
\label{algo:train}
\SetKwInOut{Input}{Input}
\SetKwInOut{Output}{Output}
\Input{Training set $\mathbb{D}=\{(\boldsymbol{o}_k,\boldsymbol{y}_k)\}_{k=1}^{|\mathbb{D}|}$; initialized model $\mathcal{M}(\boldsymbol{\theta})$; epochs $T$; batch size $B$; learning rate $\eta$}
\Output{Trained parameters $\boldsymbol{\theta}^*$}
\For{$\text{epoch}=1$ \KwTo $T$}{
  \For{each mini-batch $\mathcal{B} \subset \mathbb{D}$ }{
    \For{each $(\boldsymbol{o}_k,\boldsymbol{y}_k) \in \mathcal{B}$}{
      Build graph $\mathcal{G}=(\mathcal{V},\mathcal{E}) \leftarrow \boldsymbol{\mathcal{F}}^{(t)}, \boldsymbol{C}$\;
      
      $\{\boldsymbol{F}_{x}^{i}, \Phi(y_i)\}_{x_i \in \mathcal{V}_\mathtt{p}} \leftarrow \textit{PNFE}(\boldsymbol{I}_i, \boldsymbol{C}_i)$\;
      
      $\boldsymbol{F}_e \leftarrow \textit{ENFE}(\boldsymbol{\mathcal{F}}^{(t)}, \boldsymbol{V})$\;
      
      $\{\Psi(y_i,y_j)\}_{(i,j) \in \mathcal{E}_\mathtt{pp}} \leftarrow \textit{PPIL}(\boldsymbol{F}_{x}^{i}, \boldsymbol{F}_{x}^{j})$\;
      
      $\{\Psi(y_i,e)\}_{(i,e) \in \mathcal{E}_\mathtt{pe}} \leftarrow \textit{PEIL}(\boldsymbol{F}_{x}^{i}, \boldsymbol{F}_e)$\;
    }
    Compute $\mathcal{L}_t(\boldsymbol{\theta})$ based on Eq.~\ref{equ:total_loss}\;
    Update parameters $\boldsymbol{\theta} \leftarrow \boldsymbol{\theta} - \eta\,\nabla_{\boldsymbol{\theta}}\mathcal{L}_t(\boldsymbol{\theta})$\;
  }
}
\end{algorithm}

\subsection{Inference}
\label{sec:infer}
Given a trained model $\mathcal{M}(\boldsymbol{\theta})$ and a temporal observation window
$\boldsymbol{\mathcal{F}}^{(t)}=\{\boldsymbol{f}_{t-l+1},\dots,\boldsymbol{f}_t\}$, we first obtain the raw network outputs: individual crossing probabilities $\hat{p}_i$ from \textit{PNFE},
interaction state distributions $\hat{p}_{ij}$ from \textit{PPIL},
and environmental feasibility scores $\hat{p}_{ie}$ from \textit{PEIL}.
Let $\mathcal{Y}=\{0,1\}^{|\mathcal{V}_\mathtt{p}|}$ denote the discrete solution space.
The final inference energy $E_{\mathtt{infer}}(\boldsymbol{y})$ aggregates the base energy and our proposed consistency penalties.
At inference, we seek the configuration $\boldsymbol{y}^*$ that minimizes the total inference energy $E_\mathtt{infer}(\boldsymbol{y})$:
\begin{equation} 
E_{\mathtt{infer}}(\boldsymbol{y}) = E_{\mathtt{base}}(\boldsymbol{y}) + \lambda_1 E_{\mathtt{pp}}(\boldsymbol{y}) + \lambda_2 E_{\mathtt{pe}}(\boldsymbol{y}), 
\end{equation} where $ E_{\mathtt{base}}(\boldsymbol{y})$ aligns with the training objective in Eq.~\ref{eq:energy}.

\textbf{Consistency Energy Terms.} A key challenge during inference is the lack of ground-truth supervision, which may lead to structural contradictions between independent branch predictions. 
For instance, the \textit{PNFE} might predict a crossing intention ($\hat{y}_i=1$) for a pedestrian, while the \textit{PEIL} branch simultaneously indicates a non-supportive environment ($\hat{y}_{ie}=0$).
To address this inconsistency and enforce structural harmony, we introduce consistency energy that penalizes discrepancies between node candidate states and edge predictions:
\begin{equation}
\begin{aligned}
& E_{\mathtt{pp}}(\boldsymbol{y}) = \sum_{(i,j) \in \mathcal{E}_\mathtt{pp}} \mathbb{I} \big(\mathcal{T}(y_i, y_j) \neq \hat{y}_{ij} \big), \\
& E_{\mathtt{pe}}(\boldsymbol{y}) = \sum_{(i,e) \in \mathcal{E}_\mathtt{pe}} \mathbb{I} \big(y_i \neq \hat{y}_{ie} \big),
\end{aligned}
\end{equation}
where $\hat{y}_{ij}= \arg \max~\hat{\boldsymbol{p}}_{ij}$ and $\hat{y}_{ie} =  \mathbb{I} (\hat{p}_{ie}>0.5)$ are the hard labels derived from the frozen network, and $\mathcal{T}(\cdot)$ is the same mapping function defined in Eq.~\ref{eq:edge_pp}.
The consistency terms are introduced only at inference rather than during training for two reasons: (1) during training, ground-truth labels provide direct supervision that already implicitly encourages consistent predictions, making additional penalties redundant; 
(2) the consistency terms rely on discrete, non-differentiable indicator functions over frozen network outputs, which would break 
the gradient flow if applied during training. They therefore serve as a post-hoc structural correction mechanism activated only at inference.
By incorporating these constraints into the global energy, \model\ can effectively rectify local errors through global energy minimization.

\textbf{Unary-Seeded Simulated Annealing (U-SSA).}
Finding $\boldsymbol{y}^* = \arg\min_{\boldsymbol{y}\in\mathcal{Y}} E_{\mathtt{infer}}(\boldsymbol{y})$ is a combinatorial optimization problem. While exhaustive search is feasible for sparse scenes ($|\mathcal{V}_\mathtt{p}|\leq 3$), the exponential growth of the solution space ($2^{|\mathcal{V}_\mathtt{p}|}$) necessitates a more efficient approach for crowded environments.
To solve this, we propose the \textit{U-SSA} strategy. Unlike standard SA~\cite{guilmeau2021simulated} which typically employs a random initialization, U-SSA leverages the high-confidence priors from the unary branch. We initialize the solution with a seed configuration $\boldsymbol{y}^{(0)}$ where $y_i^{(0)}=\mathbb{I} (\hat{p}_{i} >0.5)$.
This informed initialization places the search closer to the global optimum, enabling the algorithm to efficiently traverse the energy landscape and escape local minima with fewer iterations.

\begin{table*}[h]
\setlength{\abovecaptionskip}{0.1cm}  
\setlength{\belowcaptionskip}{-0.1cm} 
\centering
\setlength{\tabcolsep}{3.8pt}
\caption{Overall performance comparison with state-of-the-art methods on JAAD and PIE datasets.
Best results are in \textbf{bold}, second-best are \underline{underlined}. All metrics: higher is better.
Input: B = Bounding box, S = Speed, P = Pose, I = Image.
Blocks: Att = Attention.
Missing values are indicated by --. $^*$ denotes $32$ input frames (others use $16$).}
\label{tab:all_comparison} 
\renewcommand{\arraystretch}{1.3}
\resizebox{\textwidth}{!}{
\begin{tabular}{l c l l *{5}{c} *{5}{c} *{5}{c}}
\toprule
\multirow{2}{*}{Models} &\multirow{2}{*}{Year}  & \multirow{2}{*}{Blocks} &\multirow{2}{*}{Inputs} 
& \multicolumn{5}{c}{JAAD-beh} 
& \multicolumn{5}{c}{JAAD-all} 
& \multicolumn{5}{c}{PIE} \\
\cmidrule(lr){5-9} \cmidrule(lr){10-14} \cmidrule(lr){15-19}
& & & & Acc & P & R & F1 & AUC
& Acc & P& R & F1 & AUC
& Acc & P & R & F1 & AUC \\ 
\midrule
\text{Single-GRU}~\cite{kotseruba2020they} &2020 &\text{GRU} &B,S,I
&0.58  &0.67  &0.68  &0.67  &0.54
&0.65  &0.26   &0.49  &0.34  &0.59  
&0.83  &0.70  &0.64  &0.67  &0.77 \\
\text{Single-LSTM}~\cite{kotseruba2020they} &2020 &\text{LSTM} &B,S,I
&0.51  &0.63  &0.59  &0.61  &0.48
&0.78  &0.44   &0.70  &0.54  &0.75  
&0.81  &0.67  &0.61  &0.64  &0.75 \\
\text{TrouSPI-Net}~\cite{gesnouin2021trouspi} &2021 &\text{GRU+Att} &B,S,P
&0.64  &0.66   &0.91  &0.76  &0.56
&0.85  &0.57   &0.55  &0.56  &0.73   
&0.88  &0.73   &\textbf{0.89}  &0.80  &0.88 \\
PCPA~\cite{kotseruba2021benchmark} &2021 &\text{CNN+RNN+Att} &B,S,P,I
&0.58  &--   &--  &0.71  &0.50
&0.85  &--   &--  &0.68  &0.86   
&0.87  &--  &--  &0.77  &0.86\\
\text{PedGraph+$^*$}\cite{cadena2022pedestrian} &2022 &\text{GCN+CNN} &S,P,I
&0.70  &\textbf{0.77}   &0.75  &0.76  &\textbf{0.70}
&0.86  &\underline{0.58}   &0.75  &0.65  &\textbf{0.88}  
&0.89  &0.83   &0.79  &0.81  &0.90 \\
\text{FFSTP}~\cite{9743954} &2022 &\text{CNN+GRU+Att}  &B,S,P,I
&0.62  &0.65   &0.85  &0.74  &0.54
&0.83  &0.51   &0.81  &0.63  &0.82   
&0.89  &0.79   &0.81  &0.80  &0.86 \\
\text{PIT}~\cite{zhou2023pit} &2023 &\text{Att}  &P,S,I
&0.70  &0.71  &\underline{0.93}  &\textbf{0.81}  &\underline{0.65}
&\underline{0.87}  &0.54  &\textbf{0.85}  &0.66  &\underline{0.87}   
&0.91  &\underline{0.85}   &0.79  &0.82  &0.90 \\
PedCMT~\cite{chen2024pedestrian} &2024 &\text{Att} &B,S
&\underline{0.71}  &--   &--  &\underline{0.80}  &0.64
&\textbf{0.88}  &--  &--  &0.65  &0.77   
&\textbf{0.93}  &--   &--  &\textbf{0.87}  &\textbf{0.92} \\
PPCI\_{att}~\cite{alofi2024pedestrian} &2024 &\text{LSTM+Att} &B,S,P
&0.67    &--    &--    &0.77    &0.60
&0.81    &--    &--    &\textbf{0.75}    &0.78   
&0.91    &--    &--    &\underline{0.84}    &0.89 \\
PedSA~\cite{10878122} &2025 &\text{Att+ViT} &B,S,P,I
&0.67  &0.68   &0.90  &0.77  &0.60
&0.83  &0.47   &\underline{0.82}  &0.62  &0.80   
&--  &--  &--  &--  &-- \\
LSOP-Net~\cite{liu2025long}  &2025 &\text{GRU+Att+CNN} &B,S,P,I
&0.65   &0.65   &\textbf{0.98}  &0.78  &0.54
&0.85   &0.56   &0.61  &0.58  &0.75 
&0.89  &0.80   &0.82   &0.81   &0.87 \\
\rowcolor{oursbg}
\textbf{\model~(Ours)} &2026 &\text{Att+ViT+CRF}  &B,S,I
&\textbf{0.73}  &\underline{0.73}   &0.87 &0.79  &\textbf{0.70}
&\textbf{0.88}  &\textbf{0.62}   &\textbf{0.85}  &\underline{0.72}  &\underline{0.87}   
& \underline{0.92} &\textbf{0.86}   &\underline{0.88}  &\textbf{0.87}  &\underline{0.91} \\
\bottomrule
\end{tabular}}
\end{table*}

\begin{table}[t]
\centering
\footnotesize
\caption{Efficiency comparison. We report Parameters (Params) in Millions (M), CUDA Memory Usage (CMU) and Weights Memory Requirements (WMR) in MB, and Inference Time (Inference) in ms. All metrics: lower is better.}
\label{tab:efficiency}
\setlength{\tabcolsep}{2pt} 
\renewcommand{\arraystretch}{1.3}
\resizebox{\columnwidth}{!}{
\begin{tabular}{l  l  c  c  c  c}
\Xhline{0.8pt}
Models & Blocks   &Params  &CMU & WMR & Inference \\
\Xhline{0.8pt}
SingleRNN~\cite{medsker2001recurrent}    &RNN          &0.20     &17.75  &0.77  &0.14 $\pm$ 0.03 \\
HierarchicalRNN~\cite{shi2018multi}      &RNN          &1.06     &25.73  &4.05  &0.61 $\pm$ 0.02 \\
C3D~\cite{tran2015learning}             &3D CNN        &78.00     &479.80   &297.55  &5.57 $\pm$ 2.10 \\
I3D~\cite{carreira2017quo}              &3D CNN     &12.29     &356.59   &46.88   &8.83 $\pm$ 2.65 \\
PedCMT~\cite{chen2024pedestrian}        &Att &1.83      &21.16    &7.00    &3.34 $\pm$ 0.43 \\
PedGraph+~\cite{cadena2022pedestrian}   &GCN + CNN    &0.07      &81.34    &0.27    &1.74 $\pm$ 0.94  \\
\rowcolor{oursbg}
\textbf{\model~(Ours)}                  &Att+ViT+CRF    &1.25       &77.22  &4.77  &8.31  $\pm$ 4.62 \\
\Xhline{0.8pt}
\end{tabular}}
\end{table}

\section{Experiments}
\label{sec:exp}
\subsection{Experimental Setup}
\subsubsection{Datasets}  
We evaluate \model\ on JAAD~\cite{rasouli2017ICCVW} and PIE~\cite{rasouli2019pie} under standard experimental protocols.
For JAAD, we use the official split (177 / 29 / 117 clips) and report results on both JAAD-beh and JAAD-all.
For PIE, we adopt the standard 50\% / 10\% / 40\% split for training, validation, and testing.

\subsubsection{Evaluation Metrics} 
We use Accuracy (Acc), Precision (P), Recall (R), F1 score (F1) and Area Under the ROC Curve (AUC) to evaluate the performance of \model\ \cite{roy2022multi}.

\subsubsection{Implementation Details} 
All experiments are conducted on an Ubuntu server with an Intel i7-14700 CPU and two NVIDIA RTX A6000 GPUs, implemented in PyTorch 2.7.0. 
We resize pedestrian inputs to $200\times80$ (5 key frames) with a graph threshold $\delta_d=50$. 
Training uses the Adam optimizer with early stopping (patience = $10$) and imbalance handling~\cite{kotseruba2021benchmark}.
For JAAD (PIE), the batch size is 8 (16) with learning rate $1\times 10^{-4}$ ($0.5\times 10^{-4}$). The weights ($\alpha, \beta,\gamma$) are set differently for training and inference: (5,0.5,2.5) / (5.3,0.7,2.5) for JAAD and (2,1.5,1) / (2.5,1.6,1.2) for PIE.
The U-SSA uses initial temperature $\tau_0 = 1$ (cooling rate $0.95$).
At inference, $\lambda_1$ and $\lambda_2$ are set to $0.5$ and $0.3$, respectively.
All reported results are averaged over 5 runs.

\subsubsection{Baseline Algorithms}
We compare \model\ against the following baselines:
(\romannumeral 1) \textbf{Single-GRU/LSTM}~\cite{kotseruba2020they} concatenates multimodal features and uses GRU/LSTM for intention prediction;
(\romannumeral 2) \textbf{TrouSPI-Net}~\cite{gesnouin2021trouspi} encodes skeletal joints via atrous convolutions and parallel attention modules;
(\romannumeral 3) \textbf{PCPA}~\cite{kotseruba2021benchmark} processes pose, location, and speed via parallel RNN branches with 3D visual features;
(\romannumeral 4) \textbf{Pedestrian Graph+}~\cite{cadena2022pedestrian} integrates visual and speed features into a graph convolutional network;
(\romannumeral 5) \textbf{FFSTP}~\cite{9743954} fuses RGB images, semantic segmentation, and speed via attention and recurrent networks;
(\romannumeral 6) \textbf{PIT}~\cite{zhou2023pit} models pedestrian-environment-vehicle interactions via temporal fusion and self-attention;
(\romannumeral 7) \textbf{PedCMT}~\cite{chen2024pedestrian} predicts intentions from bounding boxes and speed via a cross-modal transformer;
(\romannumeral 8) \textbf{PPCI\_att}~\cite{alofi2024pedestrian} is a lightweight LSTM-attention network for intention prediction;
(\romannumeral 9) \textbf{PedSA}~\cite{10878122} combines adverse weather image enhancement with transformer-based prediction;
(\romannumeral 10) \textbf{LSOP-Net}~\cite{liu2025long} jointly models long-term and short-term observations for intention prediction.

\subsection{Experimental Results}
\subsubsection{\textbf{Overall Performance Comparison}}
We evaluate \model\ against state-of-the-art baseline methods on three real-world datasets for the pedestrian intention prediction task, using Acc, P, R, F1, and AUC, as summarized in Table~\ref{tab:all_comparison}.
From the average results, we observe that \model\ achieves strong and consistent performance across most metrics and datasets.
Specifically, \model\ demonstrates state-of-the-art performance on JAAD-all (Acc, P, R), PIE (P, F1), and JAAD-beh (Acc, AUC), while ranking among the top performers on other metrics. Notably, \model\ achieves absolute gains of \textbf{4}\% in P on JAAD-all, \textbf{2}\% in Acc on JAAD-beh, and \textbf{1}\% in P on PIE.
From the results across three datasets, we also observe relatively lower performance on specific metrics (e.g., P on JAAD-beh and F1 on JAAD-all) compared to certain baselines.
This stems from the optimization trade-off introduced by class imbalance in JAAD and PIE. Prioritizing overall robustness and global consistency inevitably leads to fluctuations between P and R across different evaluation subsets.

\subsubsection{\textbf{Qualitative Interpretability Analysis}}
\label{sec:inter}
\begin{figure*}[t]
    \centering
    \subfloat[Single pedestrian.]{ \includegraphics[height=5.8cm,width=0.30\textwidth]{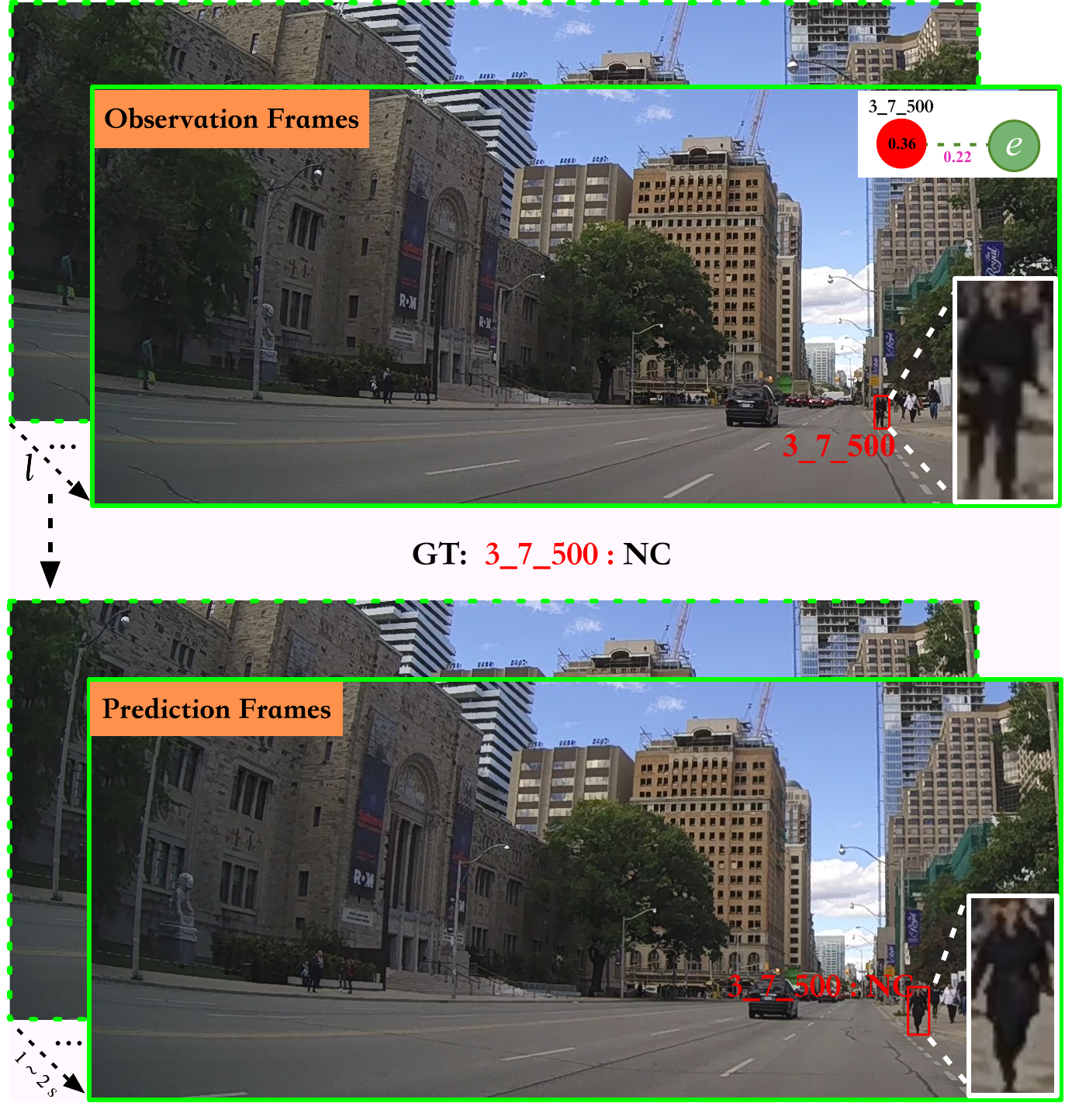}
    \label{fig:correct-q1}
    }
   \hfill
   \subfloat[Two pedestrians, same side.]{
    \includegraphics[height=5.8cm,width=0.30\textwidth]{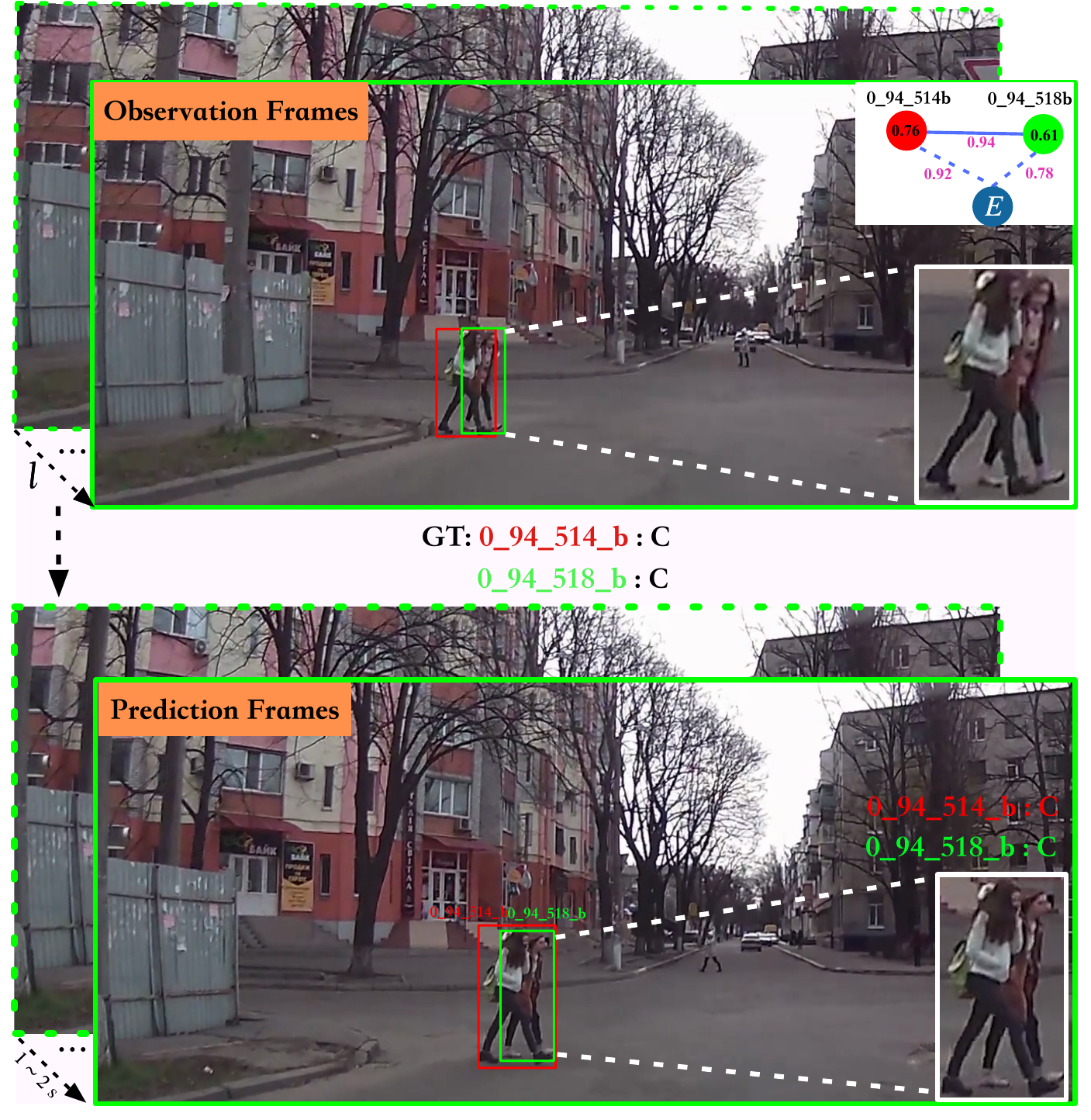}
    \label{fig:correct-q2}
   }
   \hfill
   \subfloat[Five pedestrians, opposite sides.]{
   \includegraphics[height=5.8cm,width=0.30\textwidth]{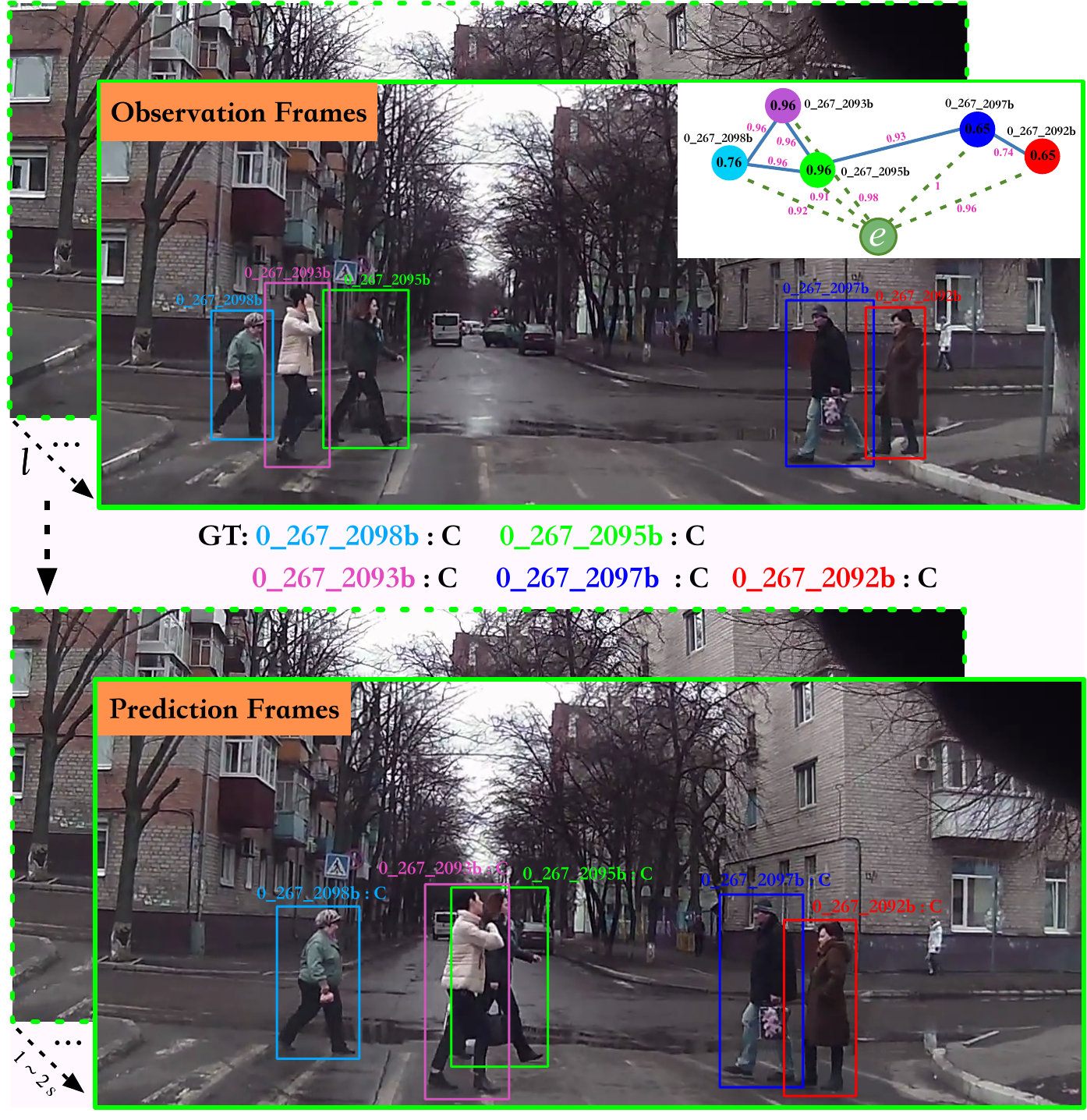}
   \label{fig:correct-q3}
   }
   \hfill
   \subfloat[Two pedestrians, same side.]{
   \includegraphics[height=5.8cm,width=0.30\textwidth]{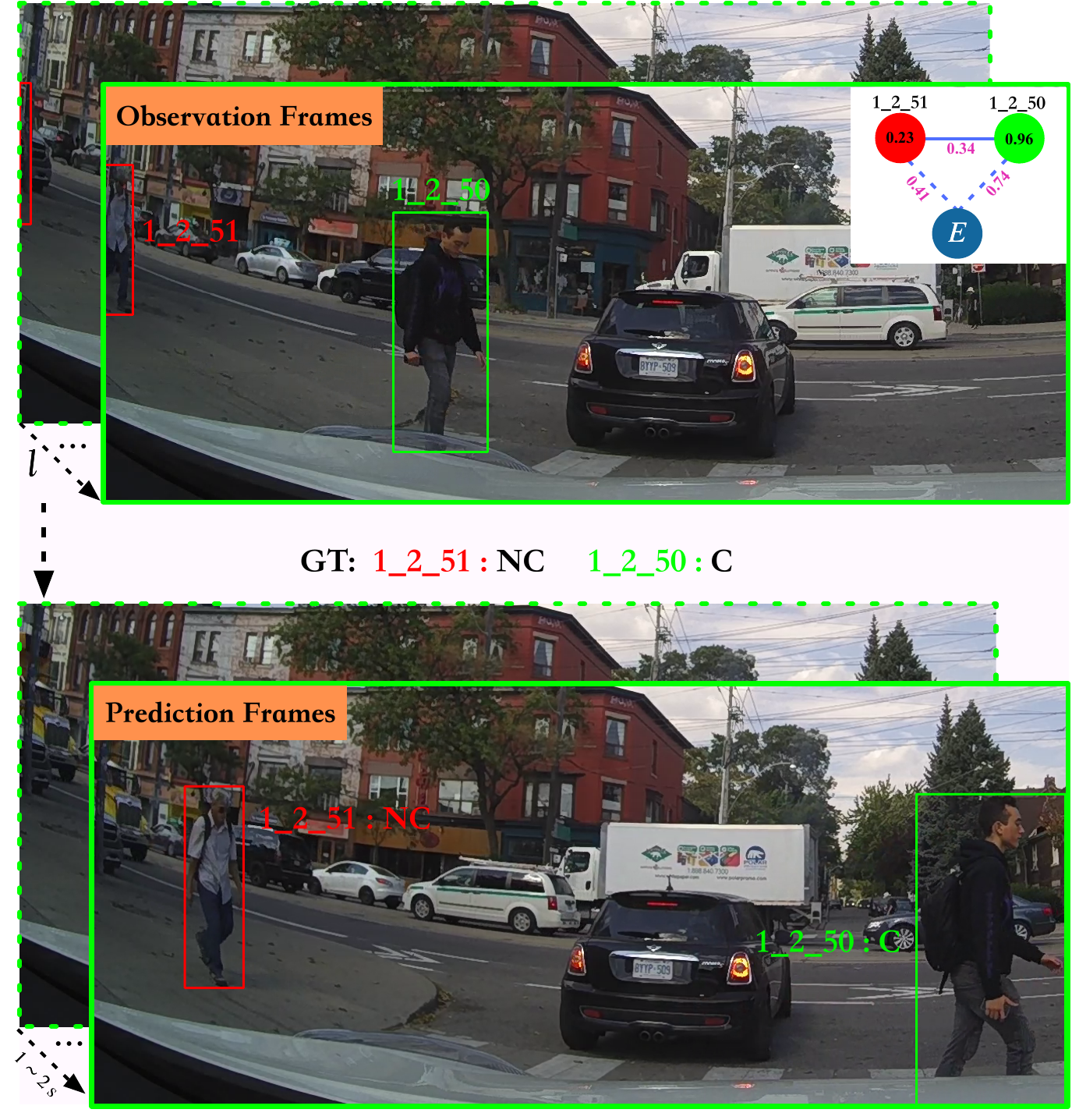}
    \label{fig:correct-q4}
   }
  \hfill
  \subfloat[Four pedestrians, same side.]{
  \includegraphics[height=5.8cm,width=0.30\textwidth]{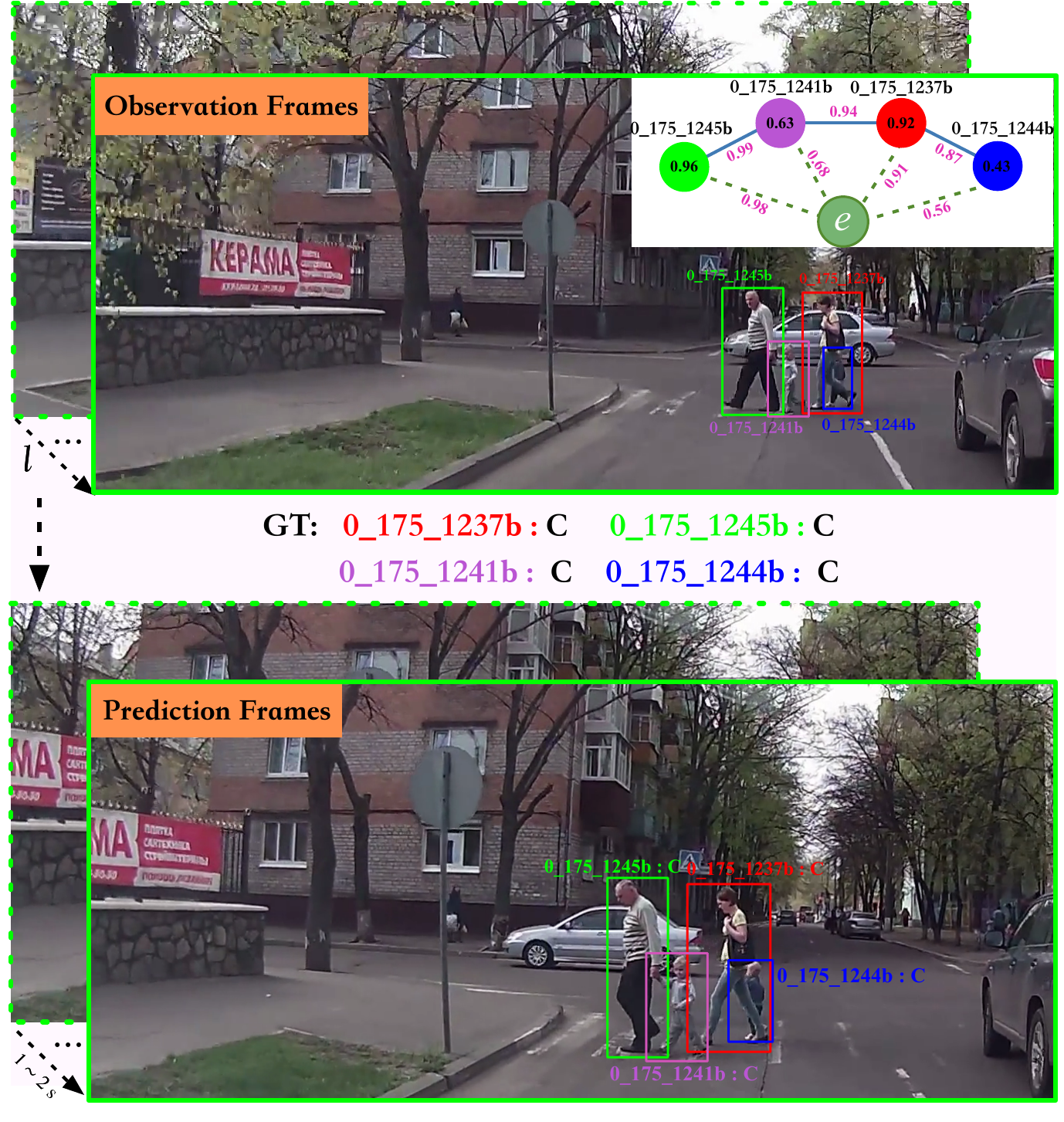}
   \label{fig:correct-q5}}
  \hfill
\subfloat[Three pedestrians, opposite sides.]{
\includegraphics[height=5.8cm,width=0.30\textwidth]{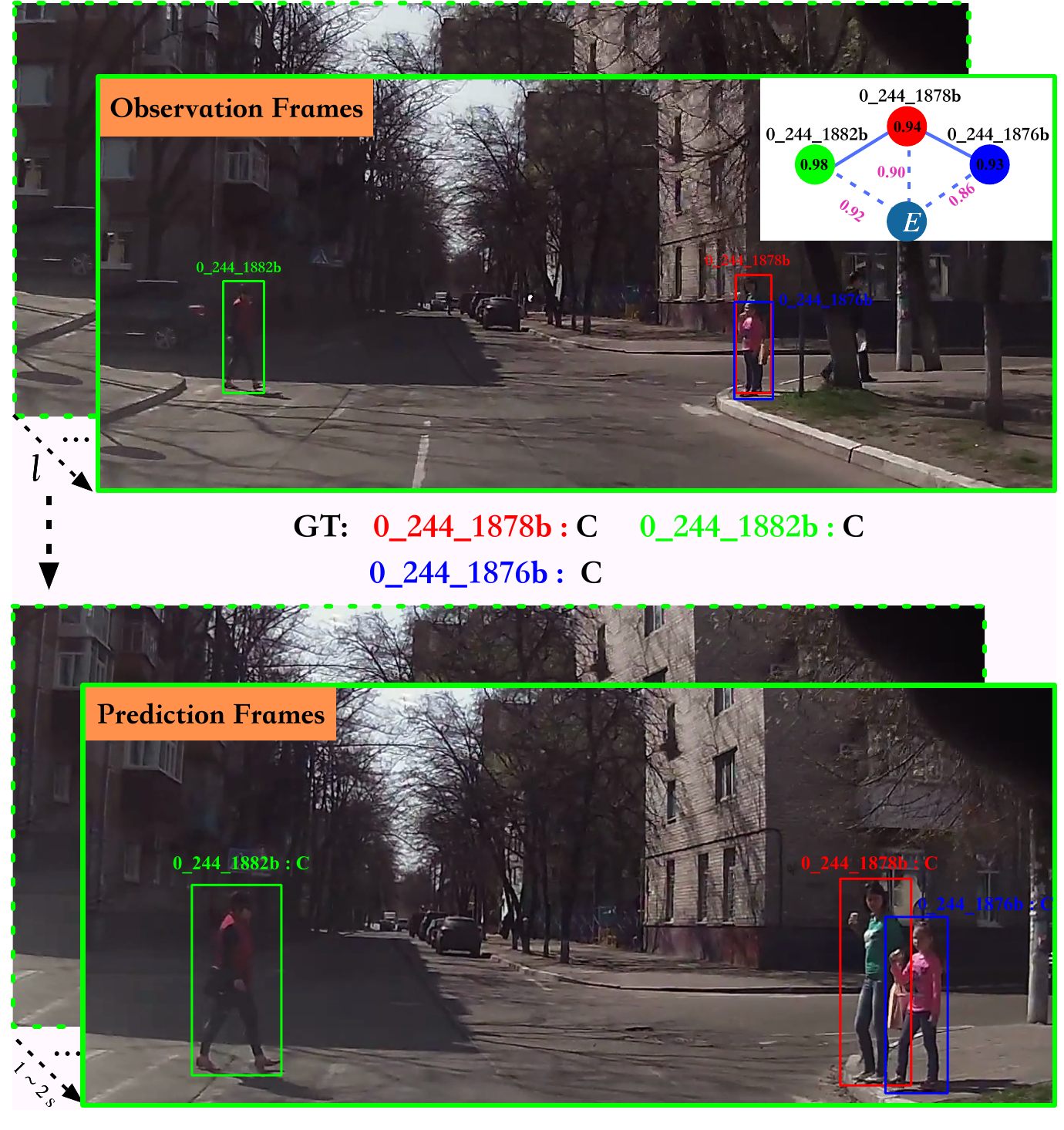}
   \label{fig:correct-q6}  
   }
  \caption{Correct qualitative results on JAAD and PIE. Scenarios span: (a) single pedestrian (PIE); (b,d) two pedestrians on the same side (JAAD/PIE); (c) five pedestrians from opposite sides (JAAD); (e) four pedestrians on the same side  (JAAD); (f) three pedestrians from opposite sides (JAAD). GT: Ground Truth.}
 \label{fig:correct}
\end{figure*}

\begin{figure*}[t]
    \centering
    \subfloat[Occlusion (JAAD).]{
        \label{fig:incorrect-q1}
        \includegraphics[height=5.6cm,width=5.4cm]{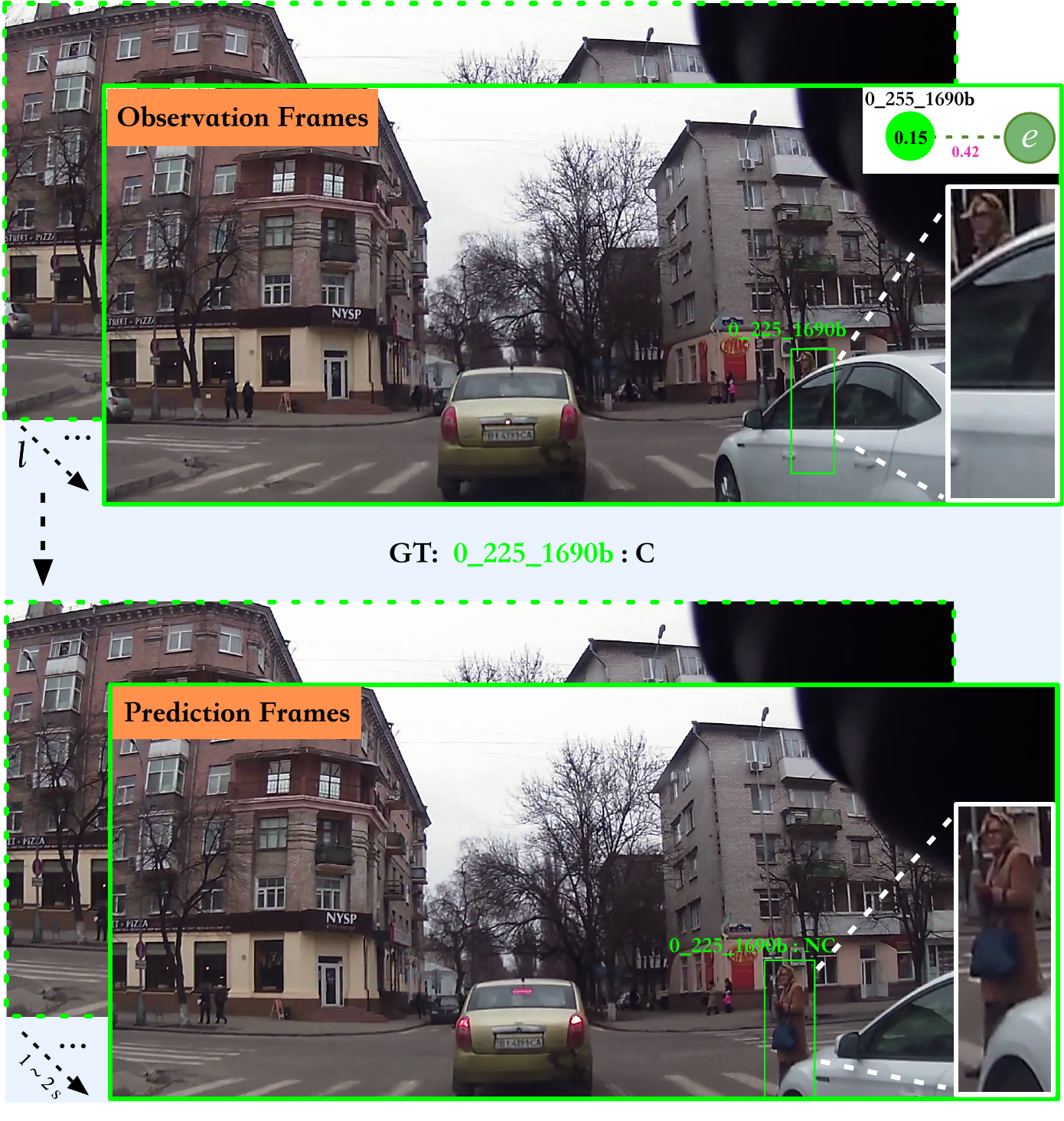}
    }
    \hfill 
    \subfloat[Adverse lighting (JAAD).]{
        \label{fig:incorrect-q3}
        \includegraphics[height=5.6cm,width=5.4cm]{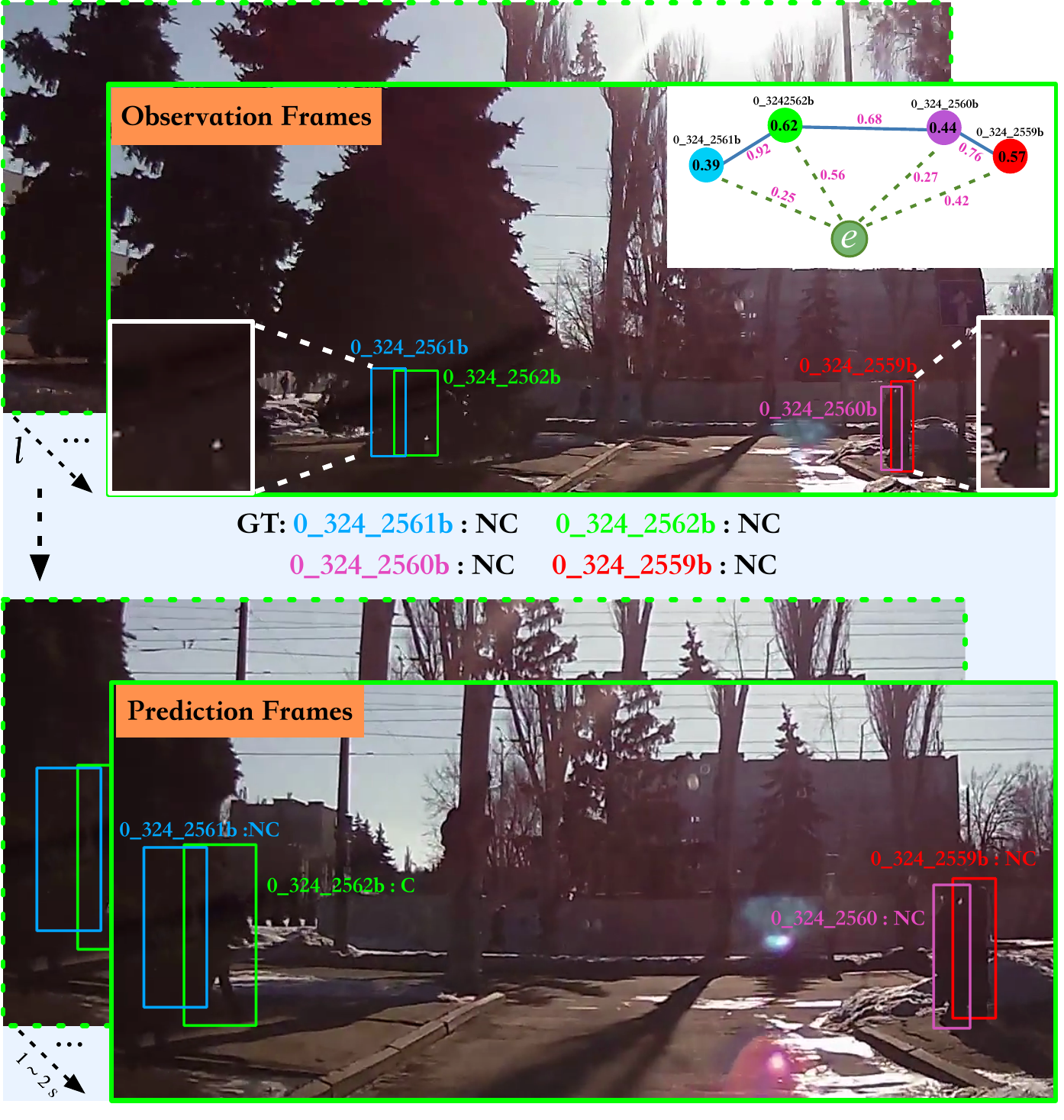}
    }
    \hfill
    \subfloat[Crowded and long distance (PIE).]{
        \label{fig:incorrect-q2}
        \includegraphics[height=5.6cm,width=5.4cm]{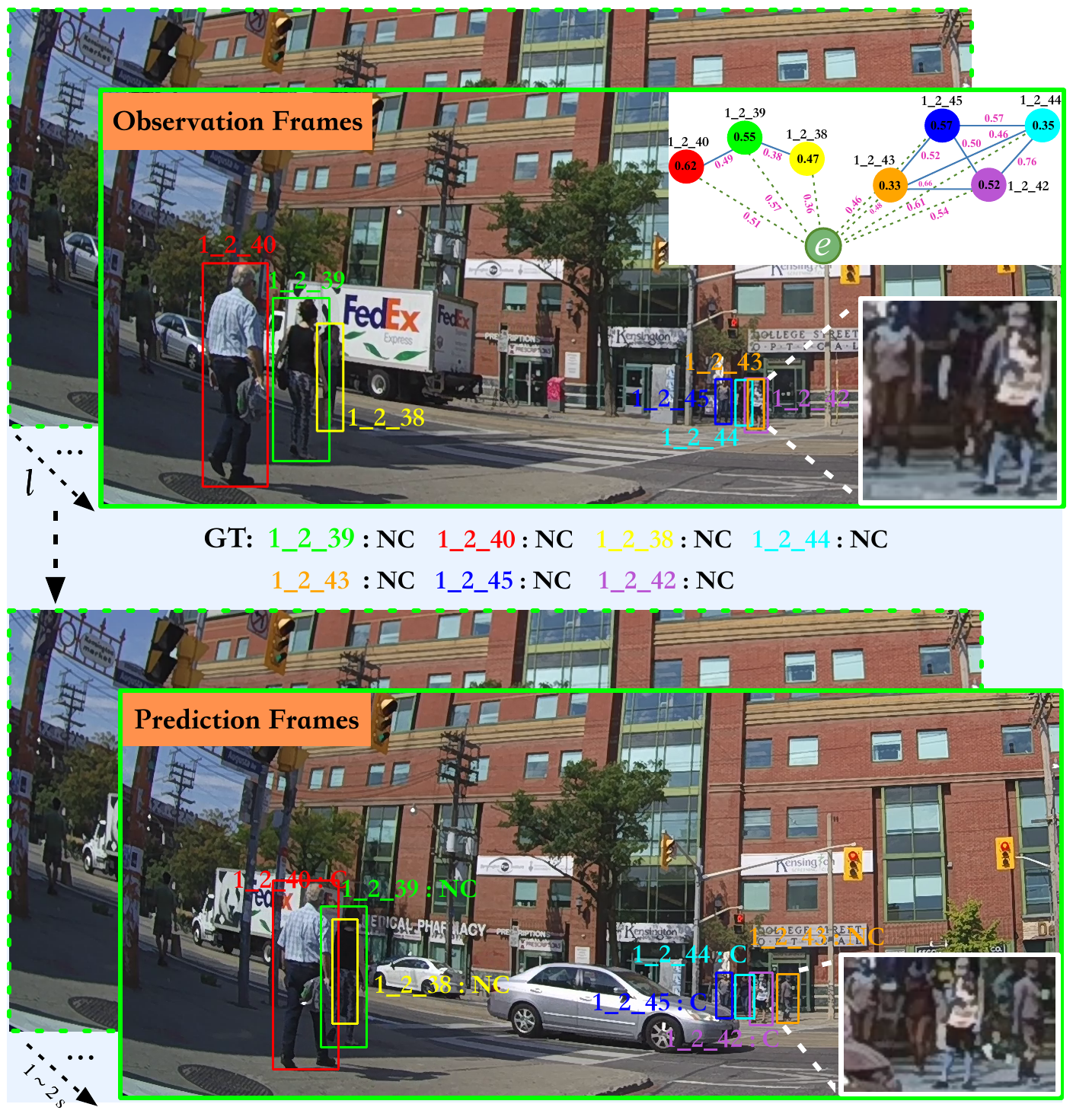}
    }
    \caption{Qualitative failure cases on JAAD and PIE. Typical error sources include (a) heavy occlusion (JAAD), (b) adverse lighting conditions (JAAD), and (c) crowded scenes with long-distance, small pedestrians (PIE).}
    \label{fig:incorrect}
\end{figure*}

For interpretability analysis, we visualize success (Fig.~\ref{fig:correct}) and failure (Fig.~\ref{fig:incorrect}) cases of \model\ in diverse scenarios from the PIE and JAAD datasets.
Each case displays observation frames (\textit{top}) and prediction frames (\textit{bottom}), with results labeled as \textbf{C} (\textit{crossing}) or \textbf{NC} (\textit{non-crossing}). Scores between pedestrians represent the \textit{PPIL}'s behavioral consistency, while those between pedestrians and the environment reflect the \textit{PEIL}'s  environmental feasibility assessment. Values within pedestrian nodes denote the intention scores from the \textit{PNFE}. For all modules, a value greater than $0.5$ signifies a positive prediction (e.g., consistency, support, or crossing intention).

Fig.~\ref{fig:correct} provides an interpretable visualization of successful prediction cases.
In the single-pedestrian scenario (Fig.~\ref{fig:correct-q1}), the prediction relies solely on the individual potential and environmental context. For instance, a score of $\textbf{0.36}$ within the pedestrian node suggests a low initial crossing probability based on pedestrian features, while the environmental support score of $\textbf{0.22}$ from \textit{PEIL} confirms the lack of external crossing cues, leading to a correct \textit{NC} prediction.
In multi-pedestrian scenarios (Figs.~\ref{fig:correct-q2}--\ref{fig:correct-q6}), the model's interpretability is further evidenced by social interaction scores. In the opposite-side scenarios (Fig.~\ref{fig:correct-q3} and Fig.~\ref{fig:correct-q6}), consistency scores of over $\textbf{0.85}$ for all P--P interactions indicate a collective crossing behavior, effectively reinforcing the model’s prediction confidence. 
Even when individual features are ambiguous---for instance, the pedestrian in the red bounding box is partially occluded by the child in the blue bounding box in Fig.~\ref{fig:correct-q6}---these pairwise potentials (P--P and P--E), representing behavioral consistency, serve as critical constraints for accurate and plausible final predictions.
Additionally, \model\ successfully captures group interactions even in complex scenarios (Fig.~\ref{fig:correct-q3}, Fig.~\ref{fig:correct-q5}, and Fig.~\ref{fig:correct-q6}).
The scores across diverse interaction types demonstrate the model's capability to simultaneously weigh individual intentions, social influence, and environmental constraints, achieving robust intention estimation.

Although \model\ performs robustly in typical scenarios, certain extreme conditions remain challenging for accurate prediction, as illustrated in Fig.~\ref{fig:incorrect}.
First, severe information loss due to heavy occlusion (Fig.~\ref{fig:incorrect-q1}) or poor illumination (Fig.~\ref{fig:incorrect-q3}) can weaken the visual cues in \textit{PNFE} and \textit{PEIL}. For example, in Fig.~\ref{fig:incorrect-q3}, the dim lighting makes it difficult for the model to extract clear motion features, leading to ambiguous intention scores.
Second, scale variation remains a challenge. When pedestrians are distant from the camera (Fig.~\ref{fig:incorrect-q2}), the subtle interaction cues between them become harder to quantify, often resulting in more conservative predictions under high uncertainty.
 Notably, Fig.~\ref{fig:incorrect-q1} represents a challenging case where dense surrounding traffic causes severe occlusion of the pedestrian, weakening the visual cues extracted by PNFE and leading \model\ to predict NC despite the ground-truth label of Crossing. This illustrates the difficulty of intention prediction under heavy occlusion in crowded scenes, where critical pedestrian features become unobservable. Addressing such cases calls for finer-grained pedestrian-level cues (e.g., gaze, gait) in future work.

\subsubsection{\textbf{Efficiency Comparison}}
This experiment evaluates the computational efficiency of \model\ in comparison with representative baseline methods, as summarized in Table~\ref{tab:efficiency}.
We adopt four commonly used efficiency metrics: Parameters (Params), CUDA Memory Usage (CMU), Weights Memory Requirements (WMR), and Inference Time. Params indicates the model complexity, while CMU and WMR reflect runtime and storage memory consumption, respectively. Inference Time measures the average latency of a single forward pass in milliseconds (ms). All reported results are averaged over 5 independent runs of 1,000 test samples each.
As shown in Table~\ref{tab:efficiency}, compared to heavy 3D CNN baselines (e.g., C3D~\cite{tran2015learning} and I3D~\cite{carreira2017quo}), \model\ dramatically reduces computational costs, lowering CMU by \textbf{84}\% and WMR by \textbf{98}\% with fewer parameters (\textbf{1.25M} vs. \textbf{78.00M}). 
Although the graph-based model (PedGraph+~\cite{cadena2022pedestrian}) is more lightweight, it suffers from inferior accuracy, as evidenced in Table~\ref{tab:all_comparison}. In contrast to the Transformer-based PedCMT~\cite{chen2024pedestrian}, \model\ reduces both Parameters and WMR by \textbf{32}\%, demonstrating superior memory efficiency. Overall, \model\ maintains a competitive inference latency of $8.31$ ms. Given a standard frame rate of $30$ FPS ($\approx33.3$ ms per frame), this speed fully meets real-time requirements.
These results confirm that \model\ achieves strong performance (Table~\ref{tab:all_comparison}) and interpretability (Sec. \ref{sec:inter}) without compromising computational efficiency or deployability.

\subsubsection{\textbf{Ablation Study}}
To evaluate the contribution of each component, we conduct a systematic ablation study as reported in Table~\ref{tab:ablation}.
Our proposed \model\ comprises four key components: (\romannumeral 1) \textit{PNFE} for extracting individual pedestrian intentions, (\romannumeral 2)  \textit{PEIL} for identifying environmental feasibility for crossing behavior, (\romannumeral 3) \textit{PPIL} for capturing pairwise interactions among pedestrians, and (\romannumeral 4) the U-SSA optimization algorithm.
Specifically, we analyze the impact of these components by comparing the full \model\ \textcircled{5} with four variants (\textcircled{1}\model$^{\texttt{-Ped.}}$, \textcircled{2}\model$^{\texttt{-P-P}}$, \textcircled{3}\model$^{\texttt{-P-E}}$, and \textcircled{4}\model$^{\texttt{-U-SSA}}$), each derived by systematically removing or replacing a specific component to quantify its contribution to the overall performance.

As reported in Table~\ref{tab:ablation}, the full \model\ achieves superior performance across all datasets, validating the holistic effectiveness of the integrated framework.
Specifically, \textcircled{5} outperforms \textcircled{1} by a significant margin (an average of \textbf{6.2\%} across all metrics). This clear enhancement demonstrates the effectiveness of the \textit{PNFE} module in capturing intrinsic behavioral cues—such as gait and posture—which serve as fundamental indicators for revealing crossing intentions.
Furthermore, we compare the full model with interaction-ablated variants. Concretely, \textcircled{5} surpasses \textcircled{2} and \textcircled{3} by average margins of \textbf{6.4\%} and \textbf{5.3\%} across
all metrics, respectively. This indicates that leveraging pairwise interaction (\textit{PPIL}) and environmental constraints (\textit{PEIL}) facilitates the preservation of social consistency and contextual awareness, thereby ensuring robust predictions even when individual features are ambiguous.
Finally, comparing \textcircled{4} with \textcircled{5}, we find that replacing the simple voting strategy with our U-SSA algorithm yields a further improvement of \textbf{4.0\%} average across all metrics. 
While the gain is moderate—as U-SSA primarily benefits multi-pedestrian scenarios which are less frequent in the dataset---it verifies the necessity of global optimization in coordinating conflicting cues for complex interaction scenarios.

\begin{table*}[htbp]
\setlength{\abovecaptionskip}{0.1cm} 
\setlength{\belowcaptionskip}{-0.2cm} 
\centering 
\caption{Ablation study of different components in \model\ on JAAD and PIE datasets. Best results are highlighted in \textbf{bold}, and second-best results are \underline{underlined}. }
\label{tab:ablation}
\setlength{\tabcolsep}{3pt} 
\footnotesize
\renewcommand{\arraystretch}{1.3}
\resizebox{\textwidth}{!}{
\begin{tabular}{l|cccc|ccccc|ccccc|ccccc}
\Xhline{0.8pt}
\multirow{2}{*}{\textbf{Model Variants}} 
& \multicolumn{4}{c|}{\textbf{Components}} 
& \multicolumn{5}{c|}{\textbf{JAAD-beh}} 
& \multicolumn{5}{c|}{\textbf{JAAD-all}} 
& \multicolumn{5}{c}{\textbf{PIE}} \\
\cline{2-5} \cline{6-20}

& \makecell{Ped.} & \makecell{P--P} & \makecell{P--E} & \makecell{U--SSA} 
& Acc & P & R & F1 & AUC 
& Acc & P & R & F1 & AUC 
& Acc & P & R & F1 & AUC \\
\Xhline{0.8pt}

\textcircled{1}\text{\model\textsuperscript{-Ped.}} 
& \textcolor{lightgray}{\XSolidBrush} & \Checkmark & \Checkmark & \Checkmark 
& 0.66 & 0.68 & 0.80 & 0.74 & 0.63 
& 0.84 & 0.54 & 0.79 & 0.64 & 0.82 
& 0.87 & 0.80 & 0.81 & 0.80 & 0.85 \\

\textcircled{2}\text{\model\textsuperscript{-P--P}} 
& \Checkmark & \textcolor{lightgray}{\XSolidBrush} & \Checkmark & \Checkmark 
& \underline{0.68} & \underline{0.69} & \underline{0.83} & \underline{0.75} & 0.65 
& 0.85 & 0.59 & 0.65 & 0.62 & 0.77 
& 0.87 & 0.77 & 0.85 & 0.81 & 0.86 \\

\textcircled{3}\text{\model\textsuperscript{-P--E}} 
& \Checkmark & \Checkmark & \textcolor{lightgray}{\XSolidBrush} & \Checkmark 
& 0.65 & \underline{0.69} & 0.75 & 0.72 & 0.64 
& \underline{0.86} & \underline{0.60} & 0.72 & 0.65 & 0.81 
& \underline{0.89} & 0.80 & \textbf{0.89} & \underline{0.84} & \underline{0.89} \\

\textcircled{4}\text{\model\textsuperscript{-U-SSA}} 
& \Checkmark & \Checkmark & \Checkmark & \textcolor{lightgray}{\XSolidBrush} 
& \underline{0.68} & \textbf{0.73} & 0.74 & 0.73 & \underline{0.67} 
& 0.85 & 0.56 & \textbf{0.86} & \underline{0.68} & \underline{0.86} 
& \underline{0.89} & \underline{0.81} & 0.84 & 0.83 & 0.87 \\

\rowcolor{oursbg}
\textcircled{5}\textbf{\model~(Ours)} 
& \Checkmark & \Checkmark & \Checkmark & \Checkmark 
& \textbf{0.73} & \textbf{0.73} & \textbf{0.87} & \textbf{0.79} & \textbf{0.70} 
& \textbf{0.88} & \textbf{0.62} & \underline{0.85} & \textbf{0.72} & \textbf{0.87} 
& \textbf{0.92} & \textbf{0.86} & \underline{0.88} & \textbf{0.87} & \textbf{0.91} \\
\Xhline{0.8pt}
\end{tabular}}
\end{table*}

\begin{figure*}[t]
    \centering
    \subfloat[Two pedestrians (JAAD).]{
   \includegraphics[width=0.31\textwidth,height=6.4cm]{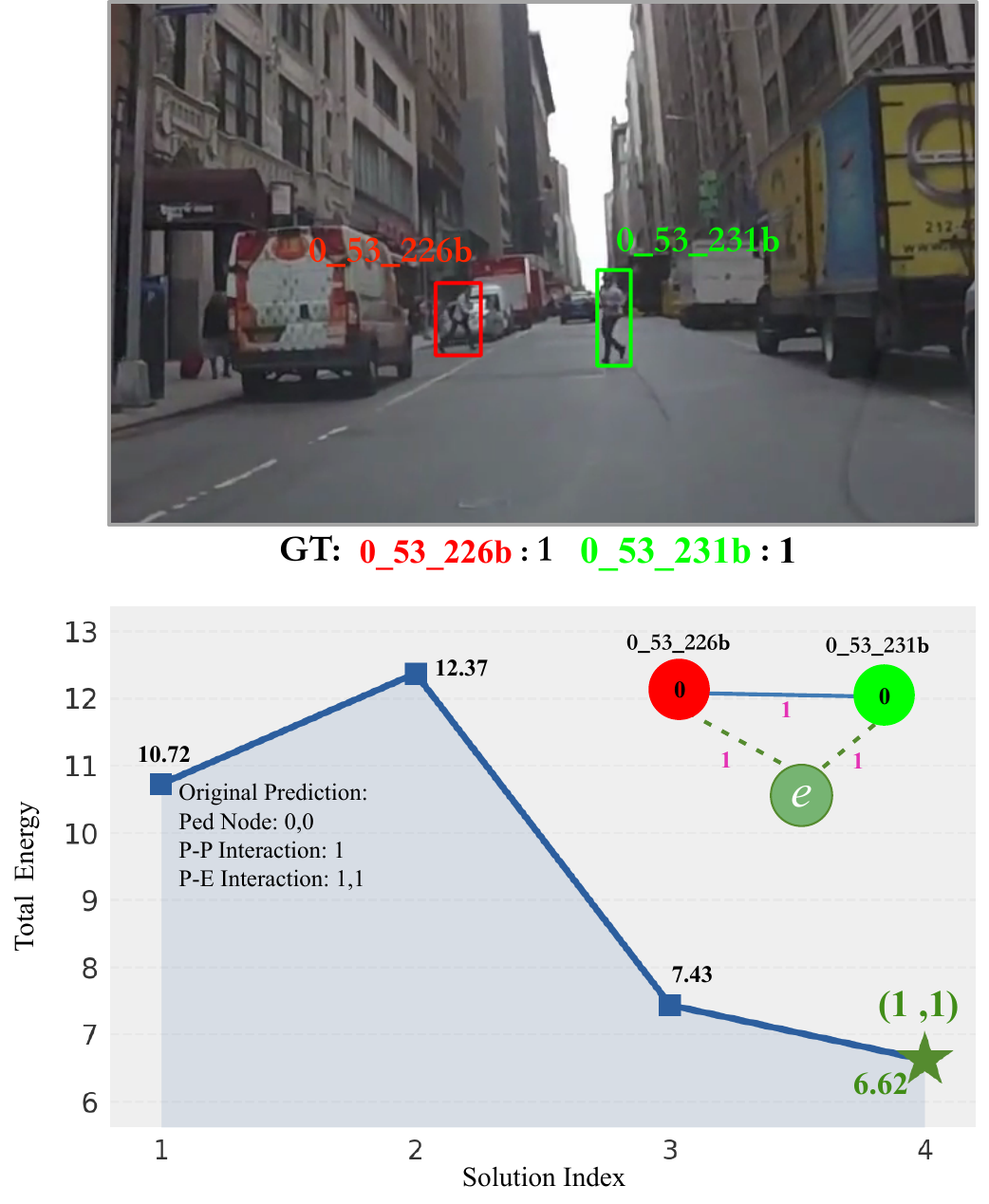}
   \label{fig:energy_2} 
   }
  \hfill
  \subfloat[Three pedestrians (JAAD).]{
   \includegraphics[width=0.31\textwidth,height=6.4cm]{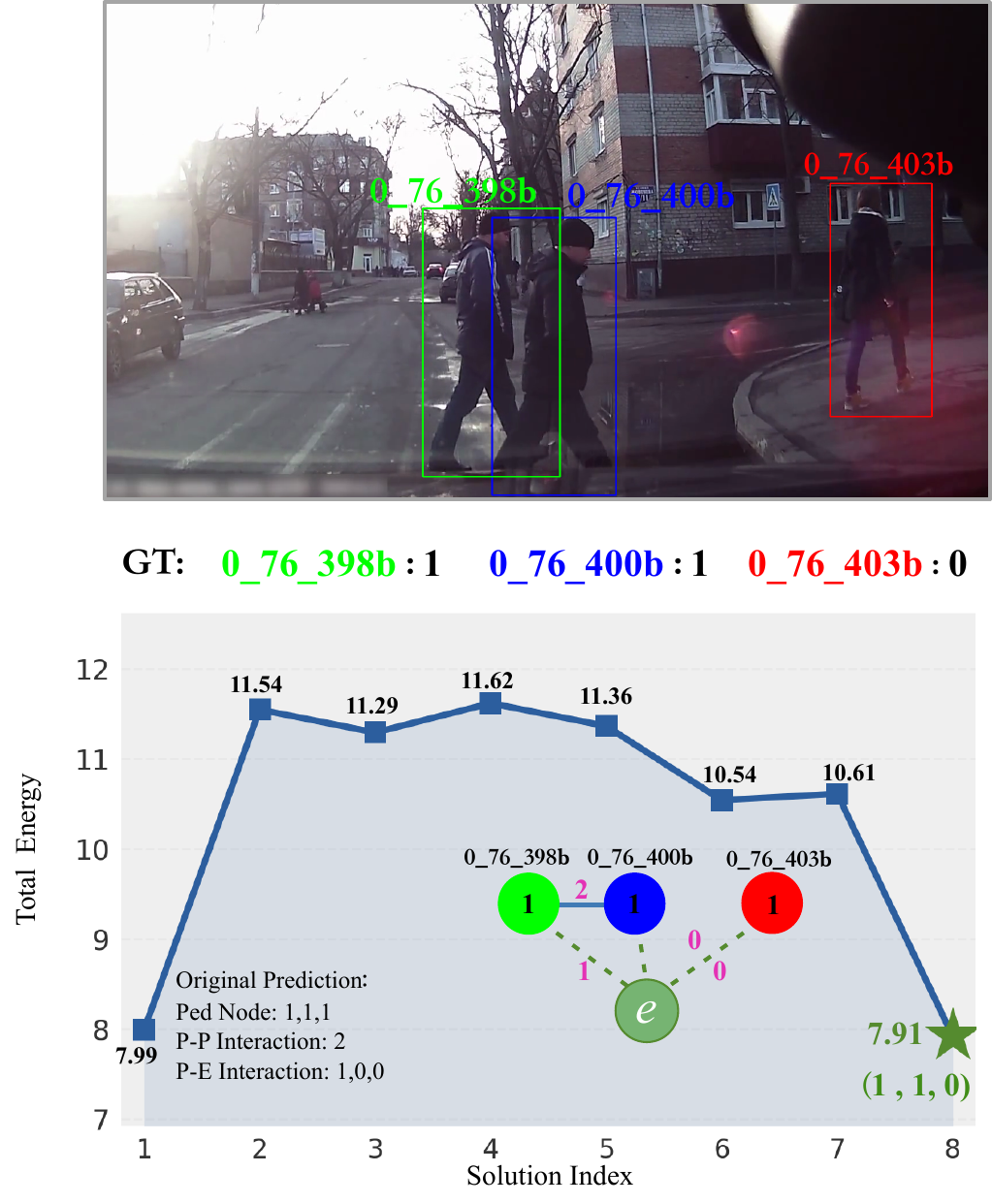}
   \label{fig:energy_3} 
   }
  \hfill
  \subfloat[Five pedestrians (PIE).]{
   \includegraphics[width=0.31\textwidth,height=6.4cm]{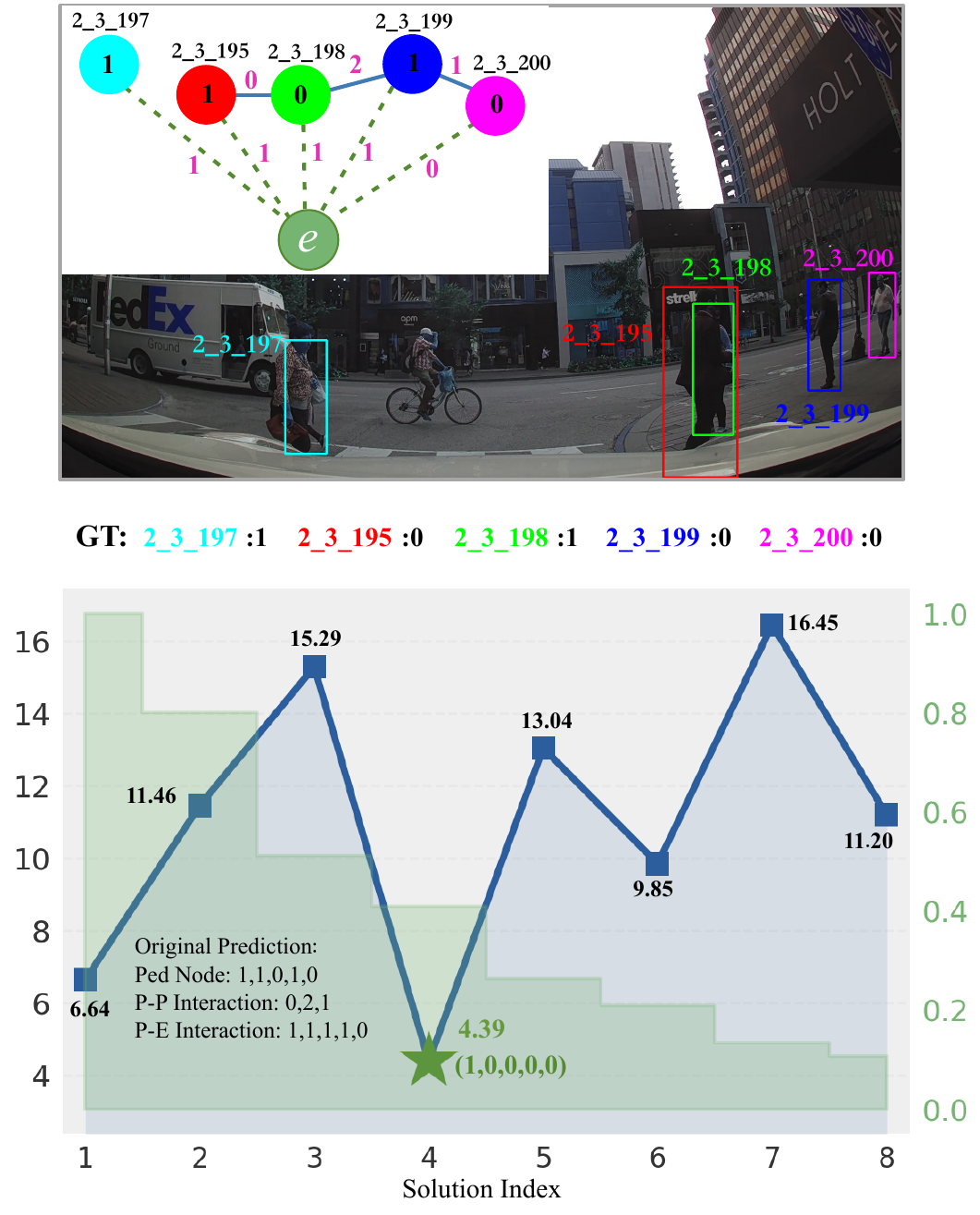}
   \label{fig:energy_5} 
   }
\caption{
Visualization of the U-SSA optimization process across scenarios with varying crowd densities. The top panels display the original scenes with GT. The bottom plots illustrate the energy minimization trajectory (blue line), where the green stars mark the optimal solutions. The interaction graphs depict the topological structure of P--E and P--P interactions. For the simpler scenarios (a) and (b), U-SSA performs an exact search due to the small solution space, whereas for the complex scenario (c), the light green bars illustrate the temperature decay.}
\label{fig:energy}
\end{figure*}

\subsubsection{\textbf{Qualitative Analysis of U-SSA Optimization}}
To provide qualitative evidence for the effectiveness of the U-SSA algorithm, we present three representative cases from the JAAD and PIE datasets in Fig.~\ref{fig:energy}. These cases illustrate the energy evolution trajectory, solution updates, and the impact of P--P and P--E interactions.
To be specific, in Fig.~\ref{fig:energy_2} (two pedestrians), the optimization starts with an initial prediction energy of $\textbf{10.72}$.
The U-SSA algorithm rapidly converges to the global optimum $\textit{(1,1)}$ with a significantly lower energy of $\textbf{6.62}$, correctly matching the GT label.
In the three-pedestrian scenario (Fig.~\ref{fig:energy_3}), the prediction is refined from the initial state $\textit{(1,1,1)}$ to $\textit{(1,1,0)}$.
This refinement is primarily driven by the P--E interaction assigning high energy penalties to the crossing intention of the pedestrian \textit{0\_76\_403b}, suggesting that the local environment (e.g., road position or signal) does not support crossing.
In the more complex five-pedestrian scenario (Fig.~\ref{fig:energy_5}), 
although the solution space expands to $2^5=32$, U-SSA identifies a low-energy solution $\textit{(1,0,0,0,0)}$ in only eight evaluations. Notably, the optimization successfully corrects the initial misclassifications for pedestrians with IDs \textit{2\_3\_195} and \textit{2\_3\_199}. While one pedestrian remains misclassified due to severe occlusion, the overall optimization demonstrates strong error-correction capability driven by the global energy minimization.
In summary, U-SSA converges to low-energy, semantically consistent solutions across varying crowd densities. By leveraging P--P and P--E interactions, it effectively corrects initial unary errors, demonstrating both computational efficiency and interpretability.

\subsubsection{\textbf{Parameter Sensitivity Analysis}}

\begin{figure*}[t]
    \centering
    \subfloat[Node coefficient $\alpha$.]{
   \includegraphics[width=6cm,height=4.3cm]{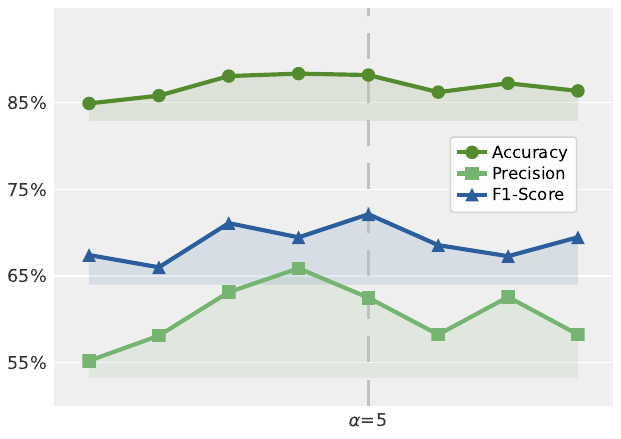}}
   \subfloat[P--P interaction coefficient $\beta$.]{
    \includegraphics[width=6cm,height=4.3cm]{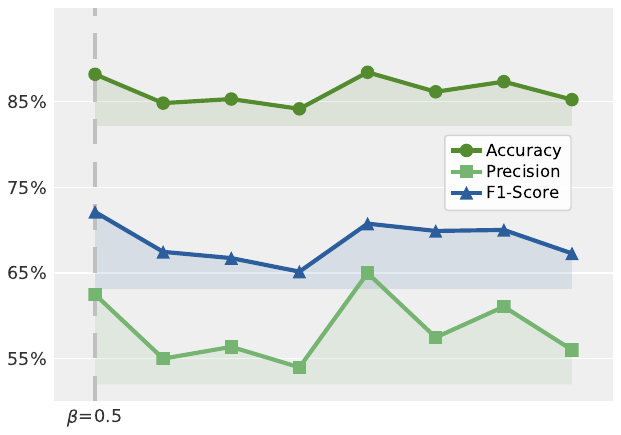}}
     \subfloat[P--E interaction coefficient $\gamma$.]{
    \includegraphics[width=6cm,height=4.3cm]{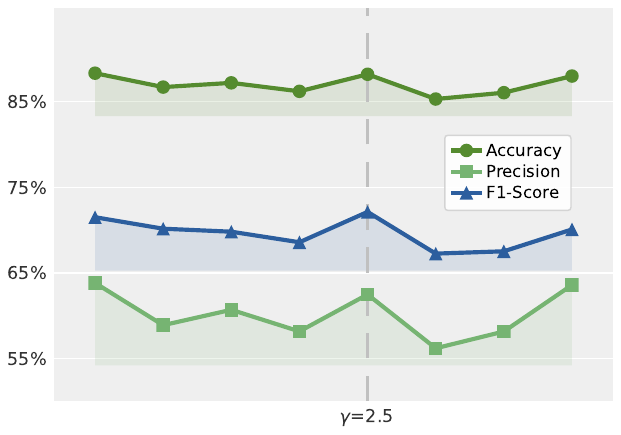}}
    \caption{Parameter sensitivity analysis of \model\ with respect to (a) node coefficient $\alpha$, (b) pedestrian–pedestrian interaction coefficient $\beta$, and (c) pedestrian–environment interaction coefficient $\gamma$ on the JAAD-all dataset. Performance is evaluated based on Accuracy, Precision, and F1-score.}  
    \label{fig:paras}
\end{figure*}

This experiment aims to evaluate the robustness of \model\ with respect to key parameters. We select the JAAD-all dataset as the representative dataset and use Accuracy, Precision, and F1-score as the main evaluation metrics.
We focus on three core parameters: the node coefficient $\alpha$, the P--P interaction coefficient $\beta$, and the P--E interaction coefficient $\gamma$. 
Specifically, we vary one parameter at a time while keeping the other two fixed, in order to analyze its influence on the model's performance. The following configurations are tested: Fixing \textbf{$\beta=0.5$} and  \textbf{$\gamma=2.5$}, we vary $\alpha \in \{1,2,3,4,5,6,7,8\}$; Fixing \textbf{$\alpha=5$} and \textbf{$\gamma=2.5$}, 
we vary $\beta \in \{0.5,1,1.5,2,2.5,3,3.5,4\}$; Fixing \textbf{$\alpha=5$} and \textbf{$\beta=0.5$}, 
we vary $\gamma \in \{0.5,1,1.5,2,2.5,3,3.5,4\}$.
The results are shown in Fig.~\ref{fig:paras}.  
Overall, \model~exhibits strong stability, with fluctuations in Accuracy and F1-score within 5\% and Precision within 7\%.  
This stability benefits from the complementary mechanisms in \model, where changes in a single parameter can be compensated by other modules through interaction modeling and global energy optimization, preventing significant performance degradation.  
In general, such robustness to parameter variations is particularly advantageous for real-world deployment in autonomous driving scenarios.  
However, noticeable performance drops are observed under certain specific parameter settings, such as $\gamma = 3$, $\beta = 2$, and $\alpha = 6$.  
This can be attributed to the high node coefficient $\alpha$ amplifying the influence of individual predictions while reducing the relative impact of interaction information; the relatively high $\beta$ may cause the P--P module to overly rely on neighborhood consistency, leading to amplified error propagation when local predictions are incorrect; and $\gamma = 3$ assigns an excessively high weight to environmental features compared to the optimal $\gamma = 2.5$, which may overemphasize environmental factors and mask pedestrian-specific behavioral cues.  Based on the experimental results, we recommend setting $\alpha$ within the range of $\left [2, 6 \right]$, $\beta$ within $\left [0.5, 2 \right]$, and $\gamma$  within $\left [1, 2.5 \right]$ in practical applications,
so as to maintain high stability while ensuring competitive model performance.

\section{Conclusion}
\label{sec:con}
In this paper, we present \model, an \textbf{E}nergy-based \textbf{S}patiotemporal \textbf{I}nteraction-\textbf{A}ware framework designed to enhance the robustness and interpretability of pedestrian intention prediction.
To address the limitations of inadequate interaction modeling, limited interpretability, and global inconsistency, we reformulate intention prediction as a structured energy minimization problem within a CRF framework.
Individual intentions are captured via unary potentials, while pedestrian--pedestrian and pedestrian--environment interactions are explicitly encoded as pairwise potentials. These are integrated into a unified global energy function to enforce scene-level consistency.
To resolve logical contradictions during inference without ground-truth supervision, we further introduce structural consistency energy terms and the Unary-Seeded Simulated Annealing (U-SSA) algorithm, which 
leverages high-confidence unary priors to efficiently converge to a globally optimal configuration.
Extensive experiments on JAAD and PIE demonstrate the superior performance of \model\ compared to baselines. 
Future work will focus on improving efficiency and extending the framework to model richer vehicle--pedestrian interactions in more complex  scenarios.


\bibliographystyle{IEEEtran} 
\bibliography{main}
\FloatBarrier
\begin{IEEEbiography}[{\includegraphics[width=1in,height=1.25in,clip,keepaspectratio]{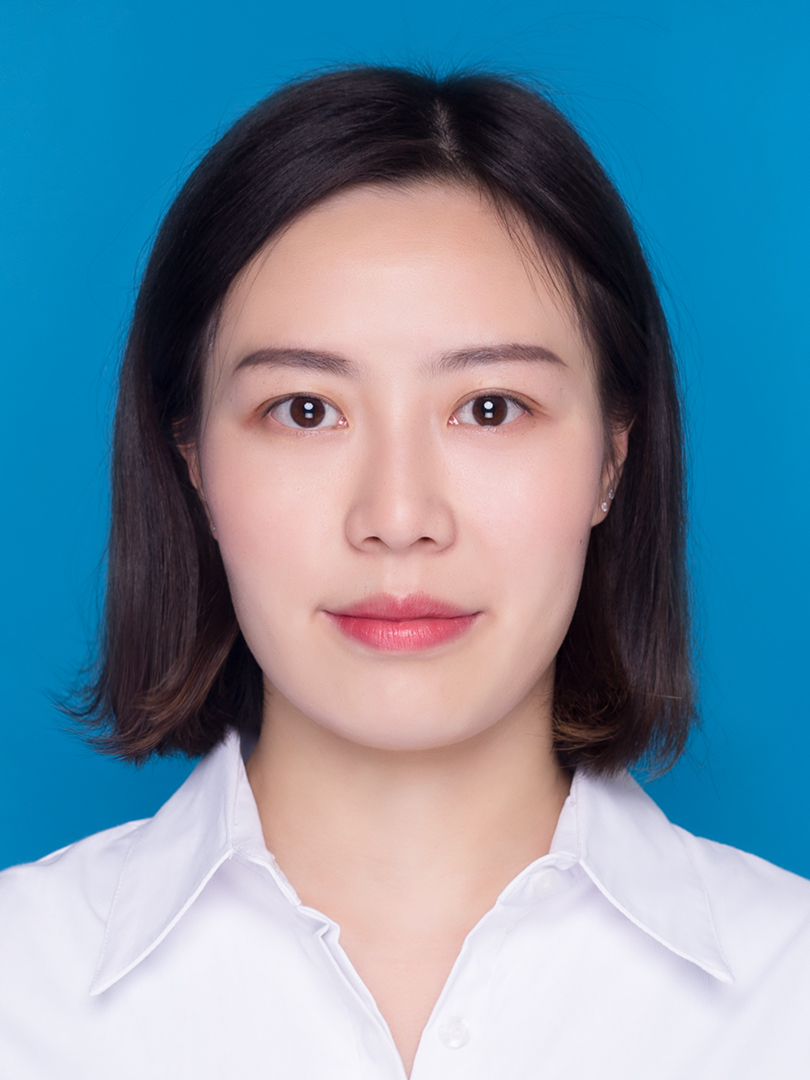}}]{Yanping Wu}
received her B.S. degree from Southwest Jiaotong University, Chengdu, China, in 2017, and her M.S. degree in 2020. From 2020 to 2023, she worked as a Machine Learning Engineer at China Merchants Bank. She is currently pursuing a Ph.D. degree in James Watt School of
Engineering at University of Glasgow, U.K. Her research interests include multi-agent interaction, multi-modal fusion, Robotics and machine learning.
\end{IEEEbiography}
\begin{IEEEbiography}[{\includegraphics[width=1in,height=1.25in,clip,keepaspectratio]{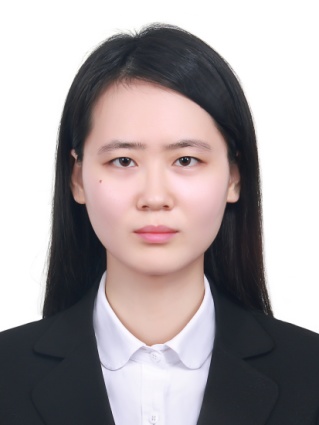}}]{Meiting Dang}
received the B.S. and the M.S. degrees from Chang’an University, China, in 2017 and 2020, respectively. She is currently working toward the Ph.D. degree in James Watt School of Engineering, University of Glasgow, U.K. Her research interests include decision-making and planning of autonomous vehicles, autonomous vehicle- pedestrian interaction modeling based on game theory, and machine learning.
\end{IEEEbiography}
\begin{IEEEbiography}[{\includegraphics[width=1in,height=1.25in,clip,keepaspectratio]{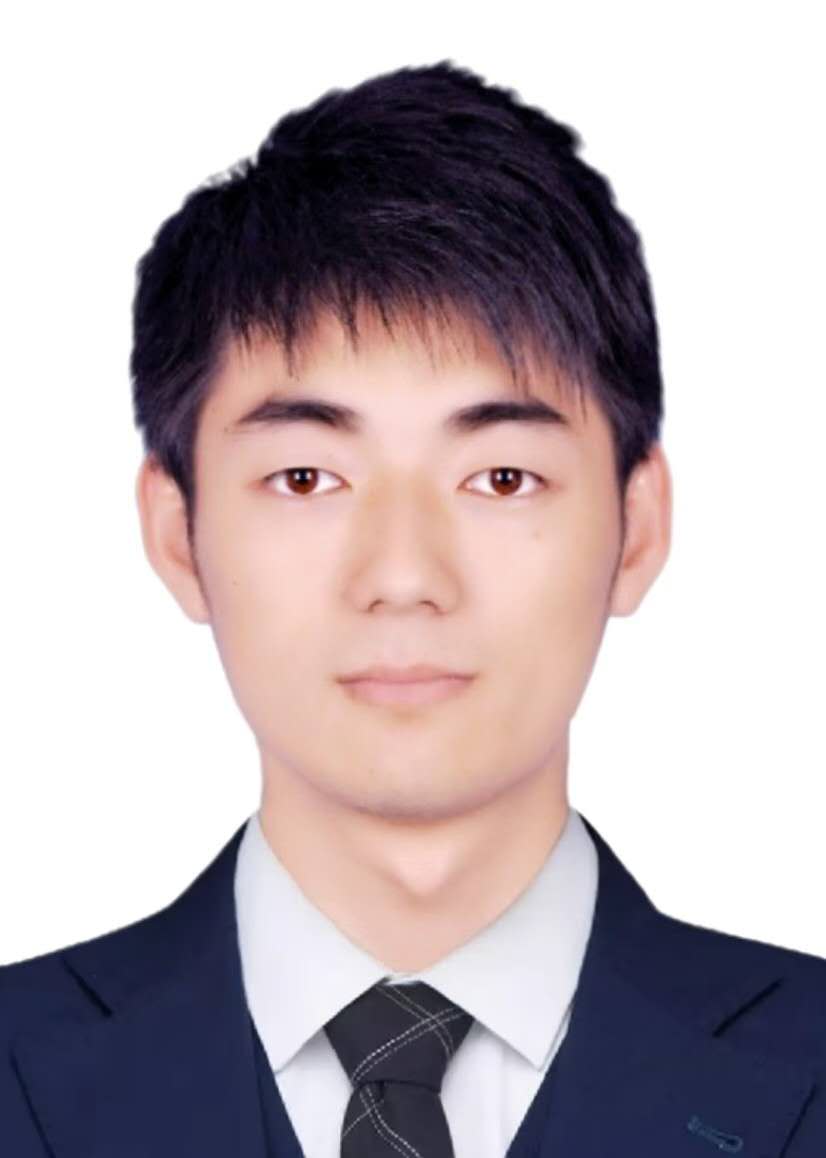}}]{Lin Wu}
received the B.S. and B.A. degrees from Wuhan University of Technology and Central China Normal University, China, respectively, in 2019, and the M.S. degree from Southeast University, China, in 2022. He is currently pursuing the Ph.D. degree with the James Watt School of Engineering, University of Glasgow, U.K. His research interests include multimodal understanding and generation, and interactive intelligent systems.
\end{IEEEbiography}

\begin{IEEEbiography}[{\includegraphics[width=1in,height=1.25in,clip,keepaspectratio]{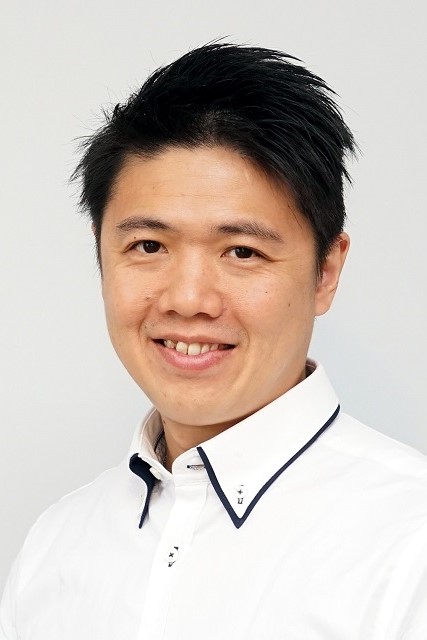}}]{Edmond S. L. Ho} is currently an Associate Professor in the School of Computing Science at the University of Glasgow, UK. He was an Associate Professor (2016-2022) in the Department of Computer and Information Sciences and Turing Network Development Award Lead (2022) at Northumbria University, Newcastle upon Tyne, UK and a Research Assistant Professor in the Department of Computer Science at Hong Kong Baptist University (2011-2016). His research interests include  Computer Vision, Biomedical Engineering, and Machine Learning.
\end{IEEEbiography}

\begin{IEEEbiography}[{\includegraphics[width=1in,height=1.25in,clip,keepaspectratio]{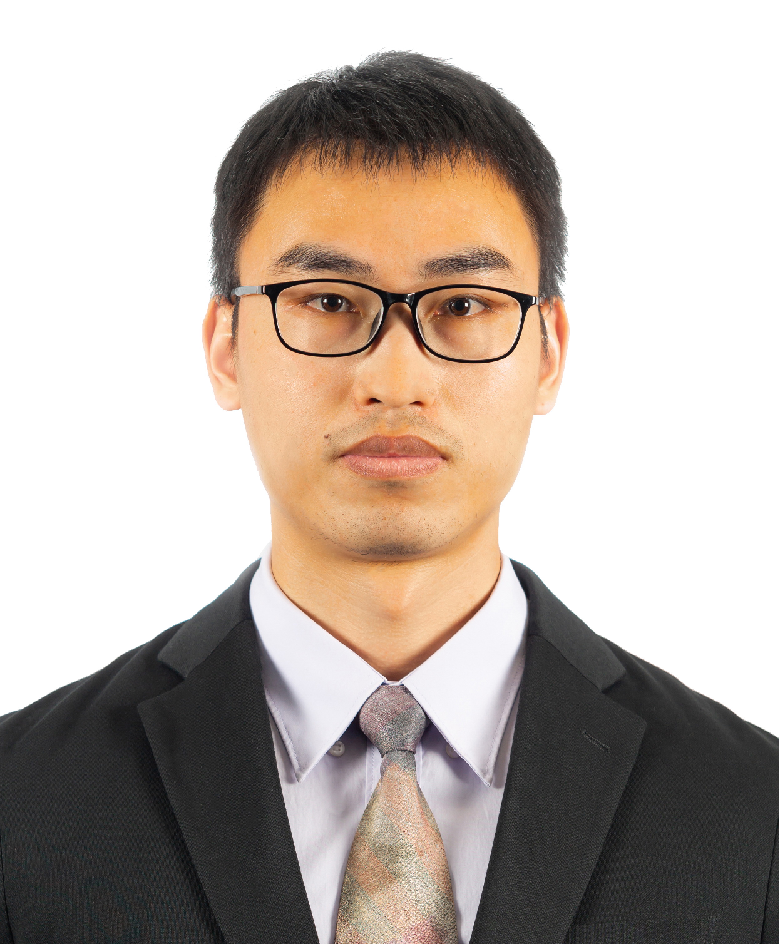}}]{Zhenghua Chen} received the B.Eng. degree in mechatronics engineering from University of Electronic Science and Technology of China (UESTC), Chengdu, China, in 2011, and Ph.D. degree in electrical and electronic engineering from Nanyang Technological University (NTU), Singapore, in 2017. Currently, he is an Associate Professor at University of Glasgow, UK. 
He serves as an Associate Editor-in-Chief of Neurocomputing and an Associate Editor for IEEE TII, IEEE TIM,IEEE T-ICPS.
His research interests include efficient AI and smart city.
\end{IEEEbiography}

\begin{IEEEbiography}[{\includegraphics[width=1in,height=1.25in,clip,keepaspectratio]{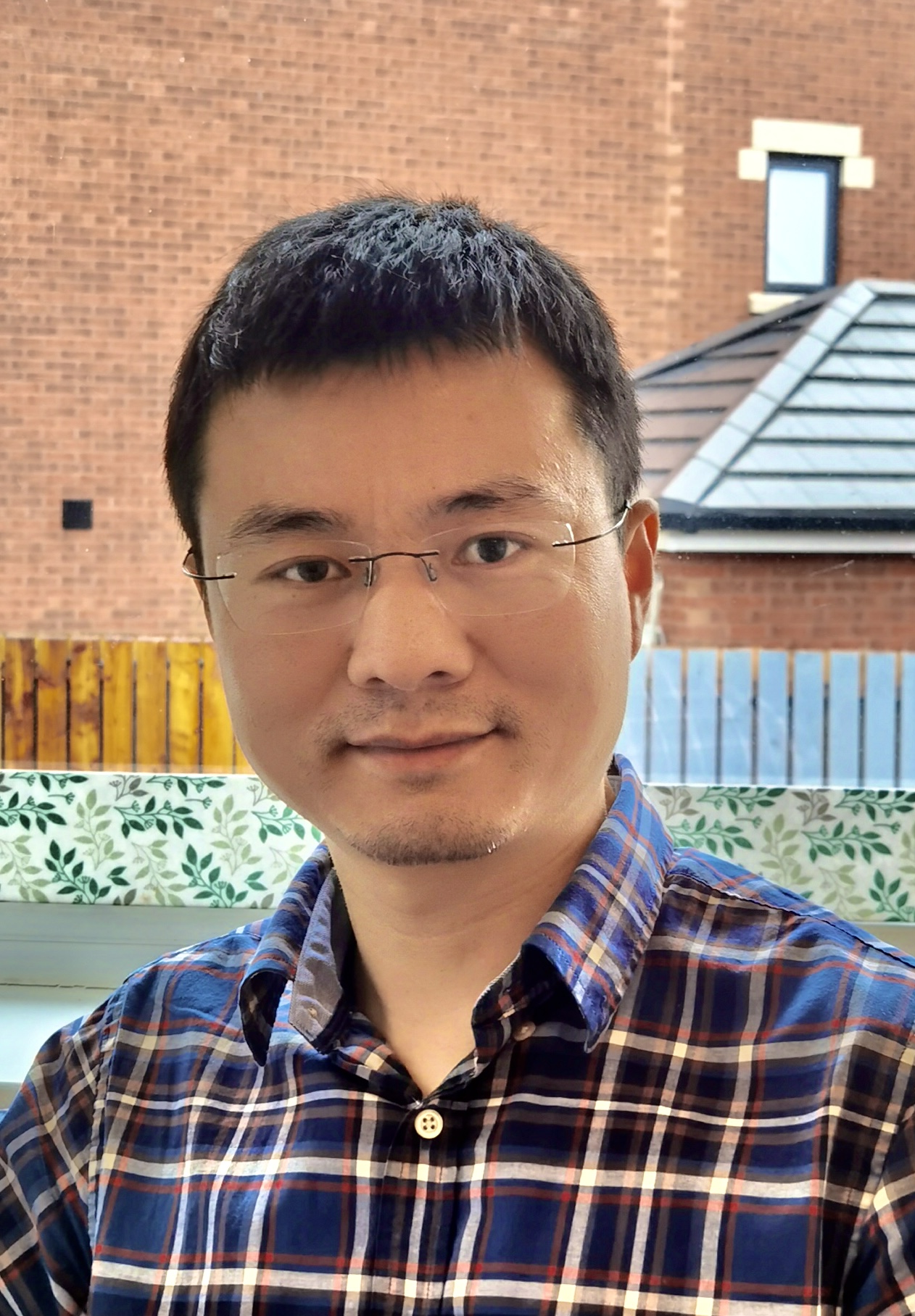}}]{Chongfeng Wei (Senior Member, IEEE)  }
 received his Ph.D. degree in mechanical engineering from the University of Birmingham in 2015. He is now a Senior Lecturer (Associate Professor) at University of Glasgow, UK. His current research interests include decision-making and control of intelligent vehicles, human-centric autonomous driving, cooperative automation, and dynamics and control of mechanical systems. He is also serving as an Associate Editor of IEEE TITS, IEEE TIV, IEEE TVT, and Frontier on Robotics and AI.
\end{IEEEbiography}

\end{document}